%% file: main.tex
\documentclass[runningheads]{llncs}

\usepackage{eccv}

\usepackage{eccvabbrv}

\usepackage{graphicx}
\usepackage{booktabs}

\usepackage{amsmath}
\usepackage{amsfonts}
\usepackage{graphicx}
\usepackage{xcolor}
\usepackage{multirow}

\renewcommand{\vec}[1]{\mathbf{#1}}
\newcommand{\T}{T}
\providecommand{\mat}[1]{\mathsf{#1}}
\usepackage{mathtools}
\usepackage{enumitem}

\newcommand{\te}[1]{\text{#1}}
\DeclarePairedDelimiterX{\norm}[1]{\lVert}{\rVert}{#1}
\DeclarePairedDelimiterX{\abs}[1]{\lvert}{\rvert}{#1}

\usepackage{siunitx}  

\usepackage{bbm}

\newcommand{\oura}{UP1PfAC}
\newcommand{\ouro}{UP2PfORI}

\DeclareMathOperator{\subdet}{subdet}
\DeclareMathOperator{\trace}{tr}

\newcommand{\Rq}{{\color{black}\mat{R}_{\text{query}}}}
\newcommand{\Rxz}{{\color{black}\mat{R}_{xz}}}
\newcommand{\Ry}{{\color{black}\mat{R}_{y}}}
\newcommand{\tq}{{\color{black}\vec{t}_{\text{query}}}}
\newcommand{\Rref}{{\color{black}\mat{R}_{\text{ref}}}}
\newcommand{\tref}{{\color{black}\vec{t}_{\text{ref}}}}
\newcommand{\pq}{{\color{black}\vec{p}_{\text{query}}}}
\newcommand{\pref}{{\color{black}\vec{p}_{\text{ref}}}}
\newcommand{\nref}{{\color{black}\vec{n}_{\text{ref}}}}
\newcommand{\tpref}{{\color{black}\tilde{\vec{p}}_{\text{ref}}}}
\newcommand{\ppref}{{\color{black}\vec{p}'_{\text{ref}}}}
\newcommand{\aref}[1]{{\color{black}{#1}_{\text{ref}}}}
\newcommand{\sref}{{\color{black}s_{\text{ref}}}}

\newcommand{\cquery}{{\color{black}c_{\text{query}}}}
\newcommand{\squery}{{\color{black}s_{\text{query}}}}

\usepackage{standalone}
\usepackage{tikz}
\usepackage{pgfplots}
\pgfplotsset{
    table/search path={figures},
}

\input{figures/tikzdefinitions}

\pgfplotsset{compat=1.18}

\usepackage{comment}

\usepackage{hyperref}

\usepackage{orcidlink}

\begin{document}

\title{Gravity-aware partially calibrated \\absolute pose estimation
from  affine-\\ or rotation-covariant features}

\titlerunning{Gravity-aware partially calibrated absolute pose estimation}

\author{Marcus Valtonen \"{O}rnhag\inst{1}\orcidlink{0000-0001-8687-227X} \and
Alberto Jaenal\inst{2}\orcidlink{0000-0003-0434-7004} \and
Stefan Adalbj\"ornsson\inst{1}\orcidlink{0000-0003-0403-1544}}

\authorrunning{M. Valtonen \"Ornhag et al.}

\institute{Ericsson Research, Lund, Sweden
\email{\{marcus.valtonen.ornhag,stefan.adalbjornsson\}@ericsson.com}\\
\and
University of Zaragoza, Spain \\
\email{ajaenal@unizar.es}
}

\maketitle

\begin{abstract}

Inertial measurement units (IMUs) are now standard in most consumer devices, such as smartphones, drones, and extended reality (XR) headsets. By fusing visual and inertial data, localization systems gain significantly in speed and robustness compared to vision-only or IMU-only approaches. However, traditional pose estimation methods fail to utilize the local geometric information embedded in feature descriptors like SIFT.
Recent work has proved the advantages of leveraging this information for relative and absolute pose estimation, but its application to partially calibrated absolute pose estimation remains unexplored. In this paper, we derive novel constraints for joint estimation of absolute pose and focal length, making use of a gravity vector obtained from IMU data and the feature-induced local geometry, which we use to construct two efficient solvers: \oura, that operates given a single affine correspondence and \ouro, which requires two orientation-covariant features.
Unlike traditional, semi-calibrated absolute pose methods requiring four point correspondences, our solvers benefit from fewer samples and lower computational cost, simplifying robust estimation in modern RANSAC-like frameworks. We evaluate the proposed solvers against the state-of-the-art on large-scale public datasets and demonstrate that our method achieves fast and accurate localization and focal length estimation.

\keywords{Absolute pose estimation \and Structure-from-Motion \and Camera calibration}
\end{abstract}

\section{Introduction}

Future generations of mixed-reality applications demand faster and more accurate global localization to enable collaborative multi-user experiences. In fact, a rich flora of techniques from image-based localization~\cite{sattler-etal-2017}, structure-from-motion (SfM)~\cite{snavely-etal-2006,schonberger-frahm-2016}, simultaneous localization and mapping (SLAM)~\cite{orbslam3}, and visual odometry (VO)~\cite{li-etal-2020} are available in the current literature to build such large-scale pipelines, yet solutions do not always transfer directly across applications. Klein and Murray~\cite{klein-murray-2007}, for instance, proposed a method tailored to augmented reality that deviates from typical SLAM by leveraging denser maps, lower-quality features, and decoupling tracking and mapping.

At the core of these applications lies the Perspective-$n$-Point (PnP) problem: estimating camera poses from 2D image features and 3D scene points. In practice, minimal solvers integrated into random sample consensus (RANSAC)~\cite{ransac} or its variants~\cite{loransac,gcransac} deal with noisy data and outliers, hence, traditionally, camera pose hypotheses are generated from three point correspondences using the P3P algorithm, for which efficient solutions are widely available~\cite{gao-etal-2003,ke-roumeliotis-2017,persson-nordberg-2018,ding-etal-2023}.

\begin{figure}[t]
    \centering
    \includegraphics[width=0.75\linewidth]{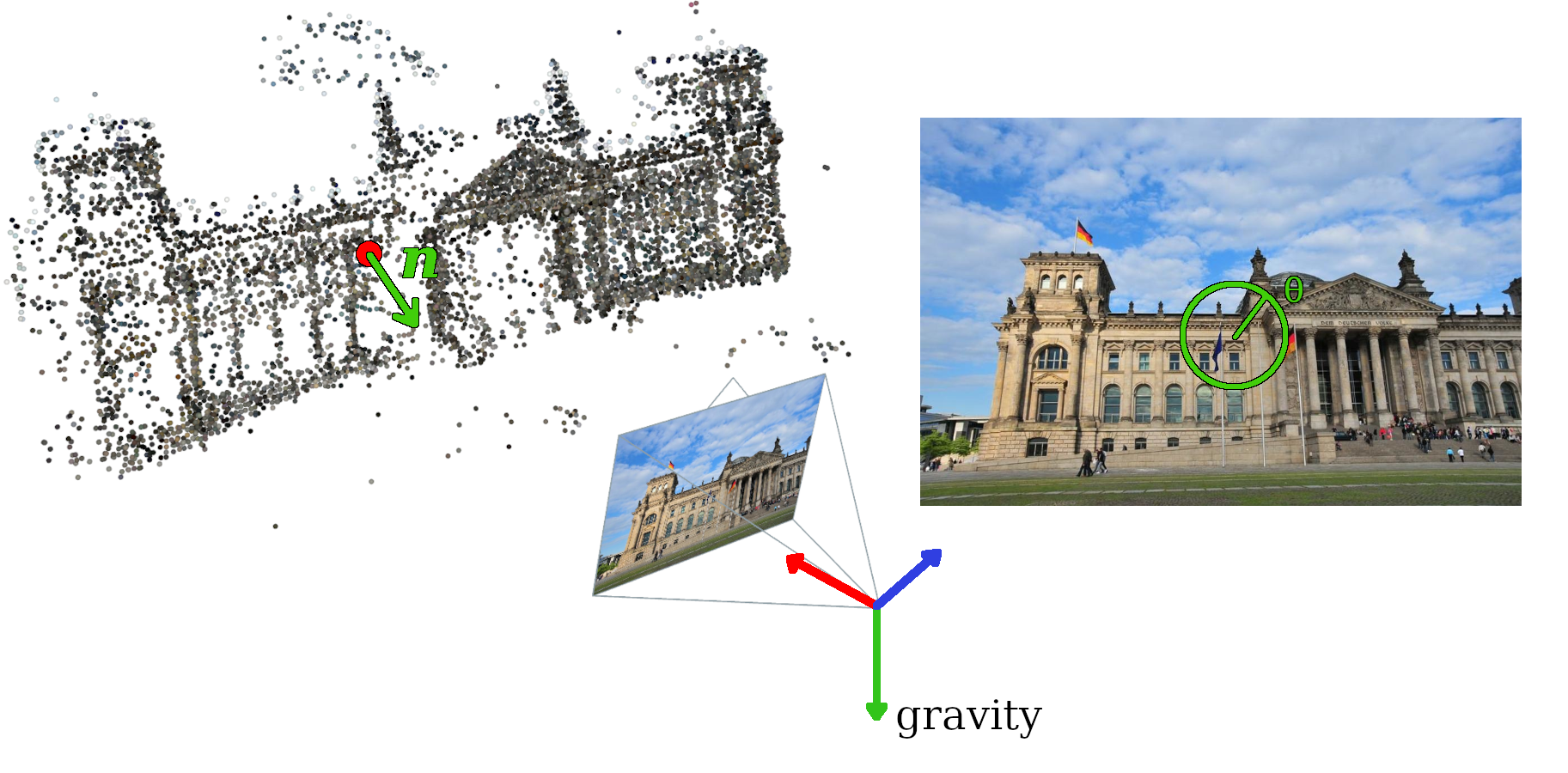}
    \vspace{-10px}
    \caption{\emph{Overview of the problem.} We propose to jointly estimate the absolute pose and focal length of a query image from a 3D reconstruction. By leveraging IMU data and the intrinsic geometric properties of local descriptors, we are able to solve the problem using a single observation 
    from an affine correspondence (\oura), or two orientation-covariant features (\ouro) to a reference image. This greatly benefits the accuracy and execution times, compared to traditional point-based methods not utilizing IMU data or feature descriptor information, which require 3.5 points or more.
    \vspace{-15
px}}
    \label{fig:first}
\end{figure}

While point-based methods have been the undisputed approach for many years, several alternative approaches that utilize affine feature descriptors were proposed in the previous decade~\cite{bentolila-francos-2014,raposo-barreto-cvpr-2016,barath-hajder-2016,barath-etal-2017-cvpr,eichhardt-2018,barath-hajder-2018,guan-etal-2019,hajder-barath-2020-icra,guan-etal-2020-cvpr,guan-etal-iccv-2021,yu-etal-2023,ding-etal-2024-cvpr}.
The improvements in this field were summarized by Barath~\etal{}~\cite{barath-etal-eccv-2020}, who
argue for its continued use and benefits. Among the drawbacks of affine-based methods is the computational overhead of feature extraction and matching. For instance, the original Affine SIFT (ASIFT)~\cite{asift} implementation reported a 13.5$\times$ increase of computational complexity  in extraction compared to regular SIFT features~\cite{sift}, and a corresponding matching step of $13.5^2 \approx 180\times$ greater complexity. With the introduction of deep learning-based descriptors, \eg,~AffNet~\cite{mishkin-etal-eccv-2018}, this gap has been reduced; however, the increased complexity remains a bottleneck.

Barath and Kukelova~\cite{barath-kukelova-2019-iccv,barath-kukelova-2022-eccv} proposed to approximate a local affine feature~$\mat{A}$ from a scale- and orientation-covariant feature, \eg{}, SIFT and SURF~\cite{surf}. They established the following constraints
\begin{align}\label{eq:aff1}
    a_1 c_{\text{ref}} + a_2 s_{\text{ref}} - q c_{\text{query}} &= 0, \\\label{eq:aff2}
    a_3 c_{\text{ref}} + a_4 s_{\text{ref}} - q s_{\text{query}} &= 0, 
\end{align}
where the relative scale is defined as~$q = q_{\text{query}}/q_{\text{ref}}$ and the orientations $\alpha_{\text{ref}}$ and $\alpha_{\text{query}}$ are used as~$c_{\text{i}} = \cos(\alpha_\text{i})$, $s_{\text{i}} = \sin(\alpha_\text{i})$. Here $a_i$ denotes the elements of~$\mat{A}$ in row-major order. If only orientation-covariant features, \eg{},~ORB~\cite{orb}, are used one may eliminate the scale by subtracting \eqref{eq:aff1} from \eqref{eq:aff2} to obtain
\begin{equation}\label{eq:ori}
c_{\text{ref}}s_{\text{query}} a_1 + s_{\text{ref}}s_{\text{query}} a_2
-
c_{\text{ref}}c_{\text{query}} a_3 - c_{\text{query}}s_{\text{ref}} a_4 = 0\;.
\end{equation}
Since their discovery, these constraints have been used to derive new minimal solvers for homography estimation~\cite{barath-kukelova-2019-iccv,ding-etal-2020b,barath-2023}, relative pose estimation~\cite{barath-kukelova-2022-eccv,valtonen-ornhag-jaenal-eccv-2024}, and calibrated absolute pose estimation~\cite{ventura-etal-cvpr-2024,hruby-etal-2024}.
In this paper, we explore applications of affine- and orientation-covariant features to improve the overall localization accuracy and computational complexity. The contributions of this paper are:

\begin{itemize}[leftmargin=2em, topsep=0.7em]
    \item Derivation of constraints for absolute pose and unknown focal length estimation, using affine or orientation-covariant feature descriptors,
    \item Two novel polynomial solvers for the upright absolute pose problem with unknown focal length using affine or orientation-covariant features
    \item A rigorous comparison to state-of-the-art methods, discussing the benefits,
    \item A practical demonstration for real-world applications with focus on visual localization with respect to a global reference system shared by other users.
\end{itemize}

\paragraph{Notation.} Following~\cite{ventura-etal-iccv-2023,ventura-etal-cvpr-2024}, we use sans-serif capital $\mat{A}$ for matrices, bold~$\vec{x}$ for vectors,
and italic lower-case $s$ for scalars. Subscripts indicate matrix and vector indexing, \eg, $\mat{R}_{1:2,1:2}$ refers to the $2 \times 2$ upper-left submatrix of~$\mat{R}$.

\section{Related work}
\paragraph{Absolute pose estimation with unknown focal length.}
The absolute pose problem with unknown focal length (PnPf) has been
well-studied in both its minimal and non-minimal configurations. Since the problem has seven degrees of freedom, 3.5 point correspondences are necessary to solve the problem, \ie{},~ignoring one of the equations obtained from the last point correspondences. Bujnak~\etal{}~\cite{bujnak-etal-2008} proposed one of the first general solvers for this case; however, the computational complexity rendered it infeasible for use in real-time applications. Zheng~\etal~\cite{zheng-etal-2014} was able to solve the problem in a fast enough way to consider it for real-time applications, which was further improved upon by Wu~\cite{wu-2015}. Later, Larsson~\etal~\cite{larsson-etal-2017} proposed a solver of a similar performance, using novel constraints for the particular setup. The overdetermined case was treated by Zheng~\etal~\cite{zheng-etal-2014} and recently by Zhou~\etal~\cite{zhou-etal-2020}.

\paragraph{Upright solvers.}
Kukelova~\etal{}~\cite{kukelova-etal-2010} derived a minimal gravity-aware two point solver solving the absolute pose problem (UP2P) also considering an upright three point solver with unknown focal length and radial distortion coefficient (UP3Pfk). Sweeney~\etal{}~\cite{sweeney-etal-2015} applied a similar approach to the UP2P solver, and demonstrated its superior performance for localization in AR applications. 
Recently, an upright solver requiring only a single scale- and orientation-covariant feature was proposed in~\cite{ventura-etal-cvpr-2024}. This solver, called UP1SIFT, was derived from~\eqref{eq:aff1}--\eqref{eq:aff2}, and the theory for local affine features, discussed in the next section.
Beyond absolute pose, gravity alignment has been leveraged for other problems such as globally optimal relative pose estimation~\cite{ding2021globally} and 3D camera registration~\cite{fragoso2020gdls}.

\paragraph{Affine correspondences and unknown focal length.}
The methods utilizing affine correspondences to estimate focal length in some way do
not treat the absolute pose problem; however, it has been studied for the relative pose problem with unknown focal length. Barath~\etal{}~\cite{barath-etal-2017-cvpr}
use two affine correspondences to solve it.
Hruby~\etal~\cite{hruby-etal-2024} showed that it is possible to use a single affine feature and estimated monocular depth. Furthermore, Ding~\etal{}~\cite{ding-etal-2024-eccv}
treated different cases of fundamental matrix estimation using relative
depths, which can be obtained from affine features,  learned scales or depths.
Recently, Yu~\etal{}~\cite{yu-etal-2026} proposed a method to solve the relative pose problem with unknown focal length and a known gravity vector by utilizing two affine features. Their approach leads to a polynomial eigenvalue problem, which can be solved efficiently and robustly, showing that correct utilization of affine features in combination with IMU data can significantly aid in simultaneous localization and camera calibration.

\section{Method}
We extend the work by Ventura~\etal{}~\cite{ventura-etal-iccv-2023,ventura-etal-cvpr-2024} by introducing an unknown focal length parameter. We propose to solve the problem using a single affine correspondence (\oura) or two orientation-covariant correspondences (\ouro) with known gravity direction,
 see~\cref{fig:first}. 

We aim to compute the unknown parameters of a query camera w.r.t. a world
coordinate system. Following the notation introduced in~\cite{ventura-etal-cvpr-2024}, the known rotational components are encoded in~$\Rxz$,
which allows for the parameterization of the full query rotation matrix $\Rq=\Ry\Rxz$. The unknown query translation is denoted~$\tq$. 
The PnP problem requires a pre-built 3D reconstruction from a set of reference camera images. As such, the known reference camera poses used to construct the map also consist of a rotation and translation denoted~$\Rref$ and $\tref$, respectively.
In this setting, the relative rotation between the reference camera and the query camera is given by~$\mat{R} =\Rq\Rref^{\T}$, and the corresponding relative translation is given by~$\vec{t} = \tq - \mat{R}\tref$.

Let $\pref$ and $\pq$ denote two corresponding inhomogeneous image point correspondences in reference and query images, respectively. Since we aim to utilize their point-based pixel coordinates to estimate the query pose and unknown focal length, we utilize the relationship between a local affine transformation and the pose of the query image relative to the reference image, as originally introduced in~\cite{ventura-etal-iccv-2023} and later modified for the world coordinate system in~\cite{ventura-etal-cvpr-2024}:
\begin{equation}\label{eq:aff}
    \mat{A}  = \frac{d}{m} \bigg(b\mat{R}_{1:2,1:2} - \ppref_{1:2}^{} \nref_{1:2}^{\T} 
    - \pq\big(b\mat{R}_{3,1:2} - \text{p}'_{\text{ref}_3} \nref_{1:2}^{\T}\big)\bigg),
\end{equation}
where $d$ is the depth of the point $\pref$ in the reference image, 
$\tpref$ is the homogeneous representation of $\pref$,
$\nref = \Rref\vec{n}$ is the normal vector transformed to the coordinate system of the reference frame, $b= \nref^T \tpref$, $\ppref = \mat{R}\tpref$, and $m = \nref^{\T} \tpref(d (\mat{R}_{3,:} \tpref) + \text{t}_3)$.
Directly working with the world coordinate system enables the usage of correspondences in different reference images, thereby increasing the number of possible samples for RANSAC. This is often utilized in modern localization frameworks, \eg{},~HLOC~\cite{sarlin2019coarse}.

\subsection{Incorporating the IMU data}
As opposed to previous work, we assume that two degrees of freedom can be reduced due to a known gravity direction, typically obtained using an IMU sensor. In this situation, the camera rotation can be decomposed as $\mat{R}=\mat{R}_y\mat{R}_{xz}$, where the known rotation is encoded in~$\mat{R}_{xz}$, aligning the remaining unknown yaw angle with the $y$-axis in the coordinate system.
Let $\theta$ denote the unknown angle of rotation about the $y$ axis of the query camera, then
\begin{equation}
    \mat{R}_y = \begin{bmatrix}
        \cos\theta & 0 & -\sin{\theta} \\
        0  & 1 & 0 \\
        \sin{\theta} & 0 & \cos{\theta}
    \end{bmatrix},
\end{equation}
The tangent half-angle parameterization $r = \tan(\theta/2)$, results in
\begin{equation}
    \cos{\theta} = \frac{1-r^2}{1+r^2}
    \quad\te{and}\quad
    \sin{\theta} = \frac{2r}{1+r^2},
\end{equation}
which is suitable for polynomial solvers. Furthermore, due to the scale ambiguity, it is possible to work with scaled rotations, \ie{} $\mat{R}^{\T}\mat{R} = s\mat{I}$, where $s\neq 0$,
further simplifying the equations. The tangent half-angle parameterization cannot represent a $180^\circ$
rotation; however, this can be addressed by applying a random rotation to the variables~\cite{hesch-2011}. 

\subsection{Novel constraints}
The unknown focal length $f_{\text{query}}$ can be introduced by substituting~$\pq\mapsto\pq/f_{\text{query}}$ in \eqref{eq:aff}, and since
the reference focal length is assumed to be known, we assume that $\tpref$ is already normalized by it. Furthermore, the affine features are affected $A \mapsto f_{\text{ref}} / f_{\text{query}}A$. 
Due to space constraints, we omit the subindex of the unknown query focal length and simply write~$f$.
This yields
{\footnotesize%
\begin{align} \label{eq:affine-constratins1}
    \frac{a_1}{f} &= \frac{d}{m}(br_{11} - p'_{\text{ref}_1} n_{\text{ref}_1} - \frac{p_{\text{query}_1}}{f} (br_{31} - p'_{\text{ref}_3} n_{\text{ref}_1} )), \\ \label{eq:affine-constratins2}
    \frac{a_2}{f} &= \frac{d}{m}(br_{12} - p'_{\text{ref}_1} n_{\text{ref}_2} - \frac{p_{\text{query}_1}}{f} (br_{32} - p'_{\text{ref}_3} n_{\text{ref}_2} )), \\ \label{eq:affine-constratins3}
    \frac{a_3}{f} &= \frac{d}{m}(br_{21} - p'_{\text{ref}_2} n_{\text{ref}_1} - \frac{p_{\text{query}_2}}{f} (br_{31} - p'_{\text{ref}_3} n_{\text{ref}_1} )), \\ \label{eq:affine-constratins4}
    \frac{a_4}{f} &= \frac{d}{m}(br_{22} - p'_{\text{ref}_2} n_{\text{ref}_2} - \frac{p_{\text{query}_2}}{f} (br_{32} - p'_{\text{ref}_3} n_{\text{ref}_2} ))\;.
\end{align}%
}%
Multiplying~\eqref{eq:affine-constratins1}--\eqref{eq:affine-constratins4} by~$mf$ yields polynomial equations.
Similarly, the orientation-covariant constraint~\eqref{eq:ori} becomes
\begin{equation}
\begin{aligned}\label{eq:ori-const}
    &\aref{c} \squery (b r_{11} - p'_{\text{ref}_1} n_{\text{ref}_1} - \frac{p_{\text{query}_1}}{f} (b r_{31} - p'_{\text{ref}_3} n_{\text{ref}_1}))\\
    &\quad+\sref \squery (b r_{12} - p'_{\text{ref}_1} n_{\text{ref}_2} - \frac{p_{\text{query}_1}}{f} (b r_{32} - p'_{\text{ref}_3} n_{\text{ref}_2}))\\
    &\quad-\aref{c} \cquery (b r_{21} - p'_{\text{ref}_2} n_{\text{ref}_1} - \frac{p_{\text{query}_2}}{f} (b r_{31} - p'_{\text{ref}_3} n_{\text{ref}_1}))\\
    &\quad-\sref \cquery (b r_{22} - p'_{\text{ref}_2} n_{\text{ref}_2} - \frac{p_{\text{query}_2}}{f} (b r_{32} - p'_{\text{ref}_3} n_{\text{ref}_2})) = 0,
\end{aligned}
\end{equation}
which are polynomial after multiplication by~$f$. Finally, the two point-based constraints are given by (multiplied after by~$f$ to become polynomial)
\begin{align} \label{eq:point-projections1}
\frac{p_{\text{query}_1}}{f} (d\mat{R}_{3,:} \tpref + t_3) - (d\mat{R}_{1,:} \tpref + t_1) &= 0, \\ \label{eq:point-projections2}
\frac{p_{\text{query}_2}}{f} (d\mat{R}_{3,:} \tpref + t_3) - (d\mat{R}_{2,:} \tpref + t_2) &= 0.
\end{align}

\subsection{Derivation of~\oura}
For the case of known gravity direction and unknown focal length with affine features, we use~\eqref{eq:affine-constratins1}--\eqref{eq:affine-constratins4} and~\eqref{eq:point-projections1}--\eqref{eq:point-projections2}, which
leaves us with a total of six equations and five unknowns: one rotational component (due to known reference direction), three translational components, and the focal length.
In general, the problem is overdetermined, and therefore, we discard one of the equations. Experimentally, we found that
discarding the last affine equation \eqref{eq:affine-constratins4} works well. This equation, however, can be utilized later, as
will be discussed.
Note that all equations~\eqref{eq:affine-constratins1}--\eqref{eq:point-projections2} are
linear in~$\tq$, hence we may write the complete system of equations in the compact form
\begin{equation}
\label{eq:base_our}
    \mat{M}(r,f)\begin{bmatrix}
        \tq \\ 1
    \end{bmatrix}
    =\vec{0},
\end{equation}
where~$\mat{M}$ is a $5\times 4$ matrix. The same methods used in~\cite{ventura-etal-iccv-2023,ventura-etal-cvpr-2024} cannot be applied, as the unknown query focal length increases the number of monomials. Instead, we note that~$\mat{M}$ must have a non-trivial nullspace;
hence, it follows that all $4\times 4$ subdeterminants of~$\mat{M}$
must vanish. Furthermore, the structure of~$\mat{M}$ can be utilized, as it may be written as
\begin{equation}
\label{eq:M}
    \mat{M} = \begin{bmatrix}
        -\xi & 0 & m_{13} & m_{14} \\
        0 & -\xi & m_{23} & m_{24} \\
        0 & 0     & m_{33} & m_{34} \\
        0 & 0     & m_{43} & m_{44} \\
        0 & 0     & m_{53} & m_{54}
    \end{bmatrix},
\end{equation}
where $\xi = f_{\text{ref}}f(1+r^2)$.
It follows that only three subdeterminants $\subdet(\mat{M})_{i,j}$ are non-trivially zero and these can be written as
\begin{equation}
    \subdet(\mat{M})_{i,j} = (1+r^2)^2 f^2 h_{i,j}(r,f),
\end{equation}
where $h_{i,j}$ are polynomials in the monomial basis $\mathcal{B}=\{r^4f, r^4, r^3f, r^3, rf, r, f, 1\}$.
As $1+r^2\neq 0$, due to our choice of parameterization, we can simply ignore these roots. Similarly, $f=0$
can be discarded as a physically improbable focal length. Additionally, only two of the three equations are linearly independent, and we may discard the second equation. The resulting system of equations is linear in~$f$, hence we may repeat the same argument as before, and write
\begin{equation}
\label{eq:rw}
    (1+r^2)^2 \bar{\mat{M}}(r)\begin{bmatrix}
        f \\ 1
    \end{bmatrix}
    =\vec{0},
\end{equation}
where $\bar{\mat{M}}$ is a $2\times 2$ matrix. Again, $\bar{\mat{M}}$ has a non-trivial nullspace.
The resulting equation~$\bar{h}(r) = \det(\bar{\mat{M}})(r)$, is quartic, hence an analytical solution is available.
In general, there are four solutions, and it can be shown that the original system of equations also has four solutions. Next, we use \eqref{eq:rw} to extract~$f$ using
backsubstitution, followed by extracting~$\tq$ from~\eqref{eq:base_our} in a similar manner. Now, the previously unused constraint~\eqref{eq:affine-constratins4} can be used to identify the correct solution, selecting the one that minimizes its residual. This guarantees that a single solution is obtained,
which is beneficial in a RANSAC-like framework, as it limits the number
of hypotheses to test against the full consensus set.

\subsection{Derivation of~\ouro}
We treat the case of two orientation-covariant features in a similar manner. Again, there are six equations and only five unknowns, and we will use only one equation of the form~\eqref{eq:ori-const}. Note that \eqref{eq:ori-const} is independent of $\tq$.
We use the point-based equations~\eqref{eq:point-projections1}--\eqref{eq:point-projections2}, which give us four equations which are linear in $\tq$,
and proceed as before, by starting from the (partial) system of equations \eqref{eq:base_our}
where~$\mat{M}$ now is a $4\times 4$ matrix. Again, it has a non-trivial nullspace, hence $\det(\mat{M}) = 0$. The structure of~$\mat{M}$ for this case is
\begin{equation}
\label{eq:M2}
    \mat{M} = \begin{bmatrix}
        -\xi_1 & 0 & m_{13} & m_{14} \\
        0 & -\xi_1 & m_{23} & m_{24} \\
        -\xi_2 & 0     & m_{33} & m_{34} \\
        0 & -\xi_2     & m_{43} & m_{44}
    \end{bmatrix}\;.
\end{equation}
where $\xi_i=f_{\text{ref},i}f(1+r^2)$. Repeating the same argument previously used, the nullspace is non-trivial, and~$\det(\mat{M})=(1+r^2)^2f^2g(r,f)$, where
\begin{equation}
\begin{aligned}
    &g(r,f) = f_{\text{ref},2}^2 (m_{13}m_{24} - m_{14}m_{23})
    + f_{\text{ref},1}^2(m_{33}m_{44}- m_{34}m_{43}) \\
    &\qquad+ f_{\text{ref},1}f_{\text{ref},2}(m_{14}m_{43} + m_{23}m_{34}  - m_{24}m_{33}- m_{13}m_{44}  )\;.
\end{aligned}
\end{equation}
Furthermore, it can be shown that $g(r,f) = (1+r^2)\bar{g}(r,f)$, due to the structure of the elements~$m_{ij}$. The resulting equation $\bar{g}(r,f) = 0$
can now be used in conjunction with one of the orientation covariant constraints~\eqref{eq:ori-const} to form a system of equations with two equations in the two remaining unknowns~$r$ and~$f$. This system is again linear in~$f$, and we may write it as
\begin{equation}
\label{eq:rw2}
    \bar{\mat{M}}(r)\begin{bmatrix}
        f \\ 1
    \end{bmatrix}
    =\vec{0},
\end{equation}
hence $\det(\bar{\mat{M}}) = 0$ gives a final equation in the unknown~$r$. This is again a quartic equation and can be solved analytically. It can be shown that the original system also has four solutions, and we may use the second (previously unused) equation~\eqref{eq:ori-const} to single out the best hypotheses
among these solutions.

\section{Experiments}
In our experiments we include the P4Pf solver by Kukelova~\etal{}~\cite{kukelova-etal-2016} and the P3.5Pf solver by Larsson~\etal{}~\cite{larsson-etal-2017}. The method by Wu~\cite{wu-2015} is reported to perform similarly to the latter and is therefore omitted. Among the gravity-aware solvers we include the recent UP1SIFT solver by Ventura~\etal{}~\cite{ventura-etal-cvpr-2024} and two solvers from~\cite{kukelova-etal-2010}: the point-based minimal two-point, UP2P, and a semi-calibrated point-based UP2.5Pf solver, which was re-implemented from the UP3Pfk solver, assuming the radial distortion parameter~$k=0$.
Furthermore, we include the affine P1AC solver~\cite{ventura-etal-iccv-2023}. Note that among these competing state-of-the-art methods, only the P4Pf, P3.5Pf and UP2.5Pf solvers estimate the unknown focal length, while the others assume this known and accounted for. Furthermore, among the methods using more than one correspondence, only the proposed \ouro~is capable of utilizing correspondences across multiple reference images simultaneously.

\subsection{Synthetic experiments}
To understand the characteristics of the proposed solvers we generate a synthetic scene by adopting the
methodology used in~\cite{eichhardt-2018,ventura-etal-iccv-2023,ventura-etal-cvpr-2024}.
Camera centers are generated for reference and query cameras, with the query positioned 2 units from the world coordinate system origin. The camera orientation is randomly generated to face the scene and the known gravity direction is obtained by decomposing it. A single reference camera is positioned at the origin with unit orientation, and random 3D points within a cube of side length~2 are generated. The unknown query focal length is chosen uniformly random in the range~$[200, 1200]$, and the 3D points are projected through the cameras to generate the image point correspondences. Following~\cite{barath-hajder-2016}, we generate local affine features by first estimating a homography corresponding to each of the 3D world points, which requires a 3D normal unit vector that is randomly drawn from a normal distribution. When the local affine transformation matrix is computed, the scales and orientations are extracted, as outlined in~\cite{ventura-etal-cvpr-2024}.
Physically improbable configurations, where 3D points are behind the cameras, the affine transformation does not have a positive determinant, or the rotation between query and reference cameras is greater than~$180^\circ$ are discarded. 
The following metrics are used:
\begin{description}[leftmargin=2em, topsep=0.7em,labelindent=2em,font=\normalfont\itshape]
    \item[Rotation error:] The standard angular distance (in degrees) between the estimated $\mat{R}_{\text{est}}$ and the ground truth rotation~$\mat{R}_{\text{gt}}$ is used, 
    \begin{equation}
        \mat{R}_{\text{err}} \coloneqq \arccos\!\left(\frac{\trace(\mat{R}_{\text{gt}}^{\T}\mat{R}_{\text{est}})-1}{2}\right),
    \end{equation}
    which can be seen as the minimum angle that aligns the rotations~\cite{Hartley2013RotationA,sattler-etal-2018}.
    \item[Position error:] The distance between the estimated camera center~$\vec{c}_{\text{est}}$ and the ground truth rotation~$\vec{c}_{\text{gt}}$, defined as $\vec{t}_{\text{err}} \coloneqq\norm{\vec{c}_{\text{gt}}-\vec{c}_{\text{est}}}$.
    \item[Focal length error:] The normalized focal length error~$f_{\text{err}} \coloneqq|f_{\text{gt}}-f_{\text{est}}|/f_{\text{gt}}$.
\end{description}

\begin{figure*}[t]
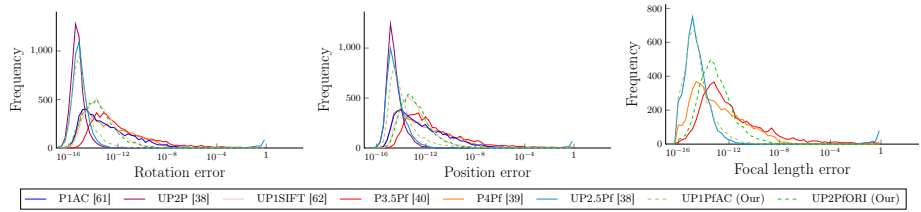

\centering
\includestandalone[width=0.325\linewidth]{figures/stability_R}
\includestandalone[width=0.325\linewidth]{figures/stability_t}
\includestandalone[width=0.325\linewidth]{figures/stability_f}
\includestandalone[width=0.965\linewidth]{figures/legend_no_markers}
\caption{\emph{Numerical stability.} Frequency histogram (\num{5000} trials) of $\log_{10}$ errors in rotation, translation, and focal length estimation across different solvers, evaluated on noise-free data. A more stable method results in a distribution shifted further to the left. Best viewed in color.}
\label{fig:stability}
\vspace{-10px}
\end{figure*}

\subsubsection{Numerical stability}
We generated \num{5000} random configurations of the synthetic environment described above, and statistics for each method are collected on them all. No noise is added to demonstrate the numerical stability of each solver, and the results are shown in~\cref{fig:stability}. 
All solvers should be considered numerically stable, with the majority of errors less than~$10^{-12}$. Among the methods estimating focal length, the proposed \oura~solver and UP2.5Pf are the most stable.

\begin{figure}[p!]
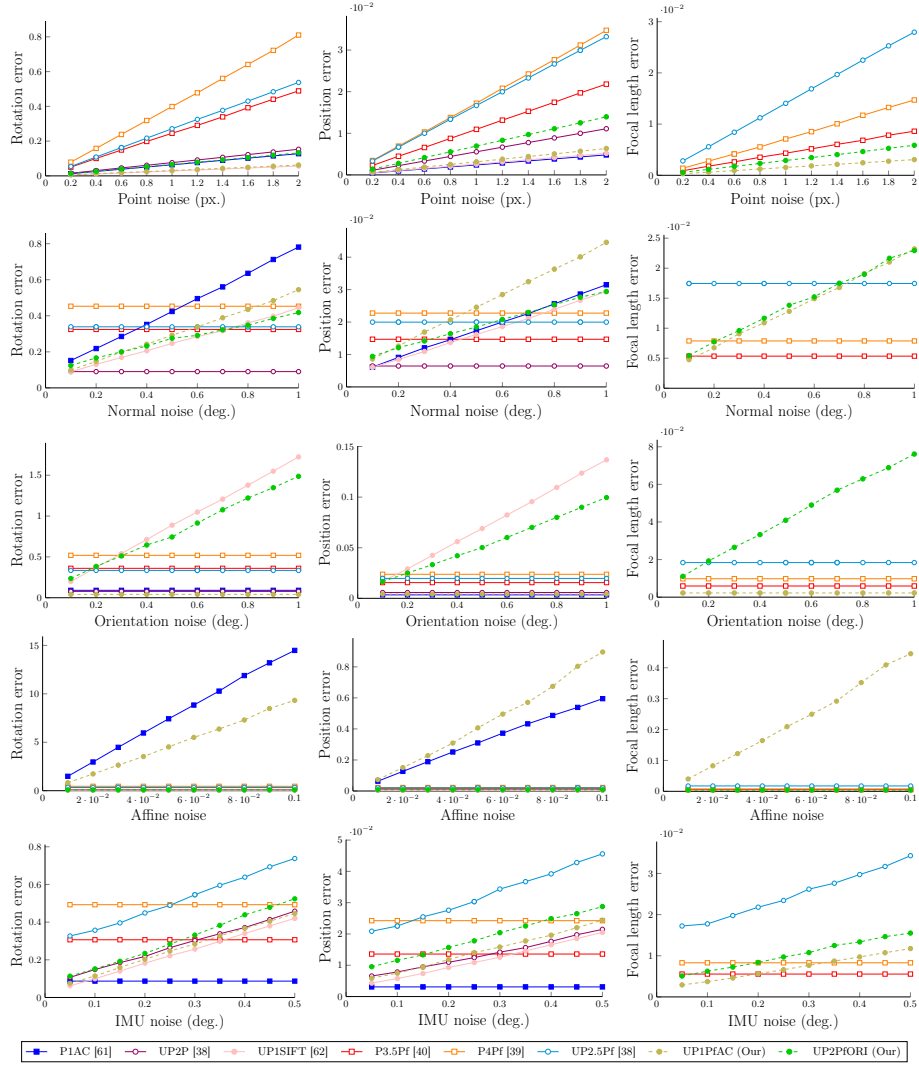

\centering
\includestandalone[width=0.325\linewidth]{figures/noise_point_R}
\includestandalone[width=0.325\linewidth]{figures/noise_point_t}
\includestandalone[width=0.325\linewidth]{figures/noise_point_f}\\
\includestandalone[width=0.325\linewidth]{figures/noise_normal_R}
\includestandalone[width=0.325\linewidth]{figures/noise_normal_t}
\includestandalone[width=0.325\linewidth]{figures/noise_normal_f}\\
\includestandalone[width=0.325\linewidth]{figures/noise_angle_R}
\includestandalone[width=0.325\linewidth]{figures/noise_angle_t}
\includestandalone[width=0.325\linewidth]{figures/noise_angle_f}\\
\includestandalone[width=0.325\linewidth]{figures/noise_aff_R}
\includestandalone[width=0.325\linewidth]{figures/noise_aff_t}
\includestandalone[width=0.325\linewidth]{figures/noise_aff_f}\\
\includestandalone[width=0.325\linewidth]{figures/noise_imu_R}
\includestandalone[width=0.325\linewidth]{figures/noise_imu_t}
\includestandalone[width=0.325\linewidth]{figures/noise_imu_f}\\
\includestandalone[width=0.965\linewidth]{figures/legend}
\caption{\emph{Noise sensitivity.} 
Median error w.r.t. image noise, normal vectors, feature orientations, affine features, and gravity vector.
All measurements report the median value over \num{1000} random test instances per noise level, where the $x$-axis shows the standard deviation of the added Gaussian noise. 
\emph{Visualization code:} circular and square markers indicate upright and non-upright orientations, respectively; filled markers denote point-based solvers, while white markers indicate alternative approaches; and dashed lines highlight our contributions. Semi-calibrated approaches are in the right column.
Best viewed in color.}
\label{fig:noise-sensitivity}
\end{figure}

\subsubsection{Sensitivity to noise}
Next, we inject Gaussian noise with zero mean and varying standard deviation to the image points to simulate matching errors naturally occurring in real localization pipelines. Image point noise is added with a constant standard deviation of 1.2~pixels; this value was chosen to represent the actual real-world
scenarios, as most modern features typically exhibit errors up to approximately 1.2~pixels~\cite{Suwanwimolkul_2021_WACV}, although methods achieving subpixel accuracy exist~\cite{kim2024keypt2subpx}.  In another experiment to simulate IMU noise, the gravity vector is perturbed with a random vector drawn from a normal distribution with zero mean and varying standard deviation. Modern
consumer devices have an IMU error of up 0.2 degrees, but high-end products may be below 0.05 degrees~\cite{kukelova-etal-2010}, which justifies the ranges of the varying noise input.

The results are shown in~\cref{fig:noise-sensitivity}, where the median value over \num{1000} random test instances is reported for each noise level and method.
Of particular importance is the focal length errors, as the proposed methods are the first among the category of affine-based and orientation-covariant solvers to incorporate unknown focal length. Regarding sensitivity to image noise, the proposed methods perform superior, and even with the addition of extra IMU noise, the proposed methods show an advantage for reasonable consumer device levels of IMU noise.

\subsubsection{Execution time}
All solvers are implemented in C++, compiled with the same optimization flags, and executed with a standard laptop equipped with an 11th Gen Intel(R) Core(TM) i5-1145G7 @ 2.60GHz. We use public implementations where available: the P4Pf~\cite{kukelova-etal-2016} and UP2P~\cite{kukelova-etal-2010} are included in PoseLib~\cite{poselib}\footnote{\scriptsize\url{https://github.com/PoseLib/PoseLib}}, and P1AC~\cite{ventura-etal-iccv-2023}\footnote{\scriptsize\url{https://github.com/jonathanventura/P1AC}} and UP1SIFT~\cite{ventura-etal-cvpr-2024}\footnote{\scriptsize\url{https://github.com/danini/absolute-pose-from-oriented-and-scaled-features}} are available directly from the authors' repositories. The P3.5Pf~\cite{larsson-etal-2017} solver was re-implemented based on the automatic Gröbner basis solver~\cite{larsson-etal-2017b} proposed by the first author. 

\begin{table}[htbp!]
    \vspace{-10px}
    \centering
    \caption{Median timings (in nanoseconds).\vspace{-8px}}
    \label{tab:execution-time}
    {\setlength{\tabcolsep}{2.1pt}\scriptsize \renewcommand{\arraystretch}{1.5}
    \begin{tabular}{ccc|ccccc}
            P1AC{\tiny\raisebox{0.8pt}{\cite{ventura-etal-iccv-2023}}} &
            UP2P{\tiny\raisebox{0.8pt}{\cite{kukelova-etal-2010}}} &
            UP1SIFT{\tiny\raisebox{0.8pt}{\cite{ventura-etal-cvpr-2024}}} &
            P3.5Pf{\tiny\raisebox{0.8pt}{\cite{larsson-etal-2017}}} &
            P4Pf{\tiny\raisebox{0.8pt}{\cite{kukelova-etal-2016}}} &
            UP2.5Pf{\tiny\raisebox{0.8pt}{\cite{kukelova-etal-2010}}} &
            \oura &
            \ouro~\\ \hline
            2740 & 484 & 1448 & 19118 & 3179 & 642 & 2586 & 2149
    \end{tabular}
    \vspace{-30px}
    }
\end{table}

\subsubsection{Integration with robust estimation frameworks}
\begin{figure}[t!]
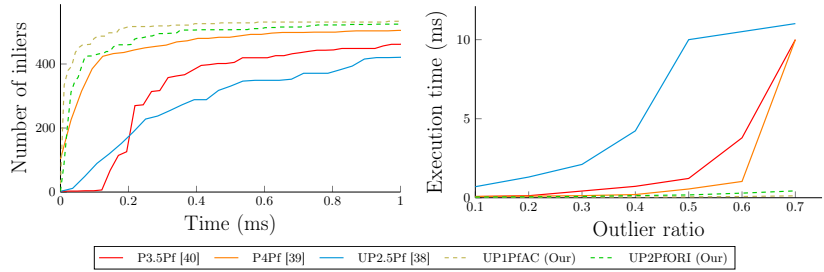

\centering
\includestandalone[width=0.44\linewidth]{figures/ransac1}
\includestandalone[width=0.44\linewidth]{figures/ransac2}
\includestandalone[width=0.70\linewidth]{figures/legend_focal}
\caption{\emph{Robust estimation.} (Left): Number of inliers found \emph{vs} execution time in a RANSAC framework with a 50~\% outlier ratio. (Right): Outlier ratio \emph{vs} total execution time to reach 80 \% of the true inlier set. Best viewed in color.}
\vspace{-8px}
\label{fig:ransac}
\end{figure}

In this section, we integrate the solvers in a standard RANSAC framework to demonstrate how they cope with outliers.
We only consider solvers where the focal length is assumed unknown, as this is the intended application of the proposed solvers.
We generate \num{1000} correspondences in each scene and use a 1.2 pixel noise and an IMU noise of 0.2 degrees. Furthermore, outliers
are injected by corrupting the data randomly. First, we consider a scenario with a 50~\% outlier ratio and compare the total number of inliers found by the respective solvers as a function of time.
To distinguish outliers from inliers a threshold of 5 pixels is used.
\cref{fig:ransac} reports the median values over 100 random problem instances. Both of the proposed solvers display advantageous performance compared to the state-of-the-art.

Furthermore, we vary the outlier ratio and report the time it takes the corresponding solvers to find an inlier set containing at least 80~\% of the true inliers. The results are shown to the right in~\cref{fig:ransac}. Here, the benefits of using more than point-based data become clearer, where the increasing exponential relationship is due to the minimal sample sizes of the corresponding solvers.

\subsection{Real data experiments}
We evaluate the proposed solvers in a visual localization scenario based on real data, where a query device must localize in the map created beforehand. For this, we use two well-known datasets: the Cambridge Landmarks~\cite{cambridgedataste} and the Aachen Day-Night v1.1~\cite{sattler-etal-2018}.
The Cambridge dataset consists of images gathered from video while walking at five different scenes in Cambridge (UK). Each scene has a train and a test split, using the first for the reconstruction and the latter as localization queries. We re-triangulate the maps by using ground-truth from VisualSfM~\cite{wu2013towards}, where the intrinsic and extrinsic calibration is obtained for each camera.
This is a relatively easy dataset because images are captured from the same device, the scene scale is medium and visual conditions are similar. We report median errors for position, rotation and focal length, as well as the median execution time.
The Aachen Dataset consists of images captured at the historical center of Aachen (Germany), a larger scale environment recorded with several different devices, which resembles more the XR scenario. It considers changes in visual appearance, making it a more challenging benchmark, especially for semi-calibrated methods. We triangulate the models based on the poses from~\cite{zhang2021reference}, and assume all query images to be upright. We follow the evaluation protocol and report the percentage of images localized within the thresholds (0.25m/2$^\circ$, 0.5m/5$^\circ$, 5.0m/10$^\circ$), as well as the median focal length estimation error.

\paragraph{Robust estimation method} We integrate our and the other solvers in the Graph-Cut-RANSAC framework~\cite{barath2018graph}, a locally optimized RANSAC alternating graph-cut and model re-fitting in the local optimization step. For the non-minimal refinement in the case of partially calibrated absolute pose estimation, we use an initial estimate from Direct Linear Transform (DLT) from the inliers of the minimal solver, which is then refined by 10 iterations of Levenberg-Marquardt.

\paragraph{Visual Localization Procedure} We followed the pipeline proposed by HLOC~\cite{sarlin2019coarse}. First, we triangulated the models from the available poses using COLMAP~\cite{schonberger-frahm-2016}. For that, we used image retrieval to recover a subset of covisible database images for each database image, using the top-20 DenseVLAD~\cite{torii201524} retrievals. Then, we extracted and matched the features from the images in the database, using two possible configurations: RootSIFT~\cite{rootsift}+Nearest Neighbor or SuperPoint~\cite{detone2018superpoint}+LightGlue~\cite{lindenberger2023lightglue}. Then, we retriangulate the model to obtain the 3D reconstructions. For localization, we retrieve the top-10 most similar database images for each query. For each of those retrieved database images, we lift the 2D-2D correspondences to 2D-3D correspondences, picking the 3D points from the SfM model related to each 2D matched keypoint in the database image. From all database images, we select the estimate with the highest number of inliers. 

In the case of SIFT, the Difference-of-Gaussians (DoG) extractor provides an estimated scale and orientation. For SuperPoint,we estimate them through S3Esti~\cite{yan2022learning}. These parameters can then be used to approximate the affine matrix $\mat{A}_i$ by $\mat{A}_i \approx \mat{S}_{q_i}\mat{R}_{\alpha_i}$, with the diagonal matrix $\mat{S}_{q,i}$ scaled uniformly by the detected scale factor $q_i$, and the rotation matrix $\mat{R}_{\alpha_i} \in \text{SO(2)}$, by the estimated angle ${\alpha_i} \in [0, 2\pi)$. Normals were estimated using the closest 200 nearest neighbors. 

\paragraph{Localization results} \cref{tab:cambridge_sp} and \cref{tab:cambridge_sift} report performance in Cambridge dataset using the SuperPoint and the SIFT pipelines, respectively. The results show that, for both features, our solvers improve the estimation errors committed by the other semi-calibrated solvers in all scenes, while also improving the time. Compared with the calibrated methods, our solvers improve performance in rotation estimation for all scenes but gets worse translation errors due to the focal length ambiguity. We observe a slight increase in terms of time, attributed to the different nonlinear refinement method used for the case of unknown focal length. Results show a better performance in general with SuperPoint, but the performance gap gets smaller when the solver is accurate enough, as in our case. \cref{tab:aachen_sp} shows performance for both SuperPoint and SIFT setups, following a similar trend than in Cambridge. As translation and rotation errors are compacted in a single metric, results show a performance degradation in case of partially calibrated methods, with our methods being better. In general, we observed that the non-linear refinement method homogenizes the final performance of the partially calibrated solvers, the main difference lying in the time.

\input{tables_cambridge}
\input{table_aachen}
\section{Discussion}
In general, we observed that the 1-point \oura \ works better than \ouro. It should be noticed that the affine constraints encode significantly more information than the orientation-covariant constraint alone. It therefore acts as a stronger geometric prior; but at a higher computational cost compared to, \eg{}, ORB. The proposed 2-point solver effectively bridges this gap, offering a compelling balance between geometric expressiveness and computational efficiency.

While modern robust estimation frameworks are equipped with numerous features, it is often unclear what truly drives their performance. Historically, research emphasized the development of fast and stable minimal solvers, but this focus has recently been reassessed~\cite{ding-etal-2024-cvpr,kocur2026minimal}. Factors such as the number of generated solutions, the choice of nonlinear refinement, and sampling strategies may be equally important---yet the speed of the minimal solver remains critical.

Our proposed method is computationally efficient, but real-world data introduces multiple error sources, including image noise, normal noise from 3D reconstruction, descriptor mismatches, and auxiliary sensor inaccuracies (\eg{}, from IMUs). Given these compounded errors, the reliability of any method is a valid concern. We demonstrate that, within a modern estimation framework, strong performance is still achievable: reasonably accurate hypotheses generated speedily, combined with effective nonlinear refinement, is sufficient to yield robust results.
With this in mind, we believe the future of minimal solvers remains bright, provided they can continue to leverage diverse sources of information, including deep learning-based descriptors, geometric priors, and auxiliary sensor data. Integrating these cues will be key to pushing the boundaries of what is possible in robust estimation and the applications thereof.

\section{Conclusions}
In this paper, we have proposed two novel solvers for the absolute pose problem with unknown focal length. To this date, these are the first solvers available in the
literature that incorporate affine- or orientation-covariant features for this problem configuration.  In order to derive the solvers, novel polynomial constraints were derived and
incorporated. Our results demonstrate that by utilizing more than point-based information extracted from the feature descriptors, we are able to significantly improve the execution speed in robust estimation frameworks compared to state-of-the-art methods, while maintaining superior or equal accuracy, both in synthetic environments and in localization tasks. Solvers of this kind pave the way for future generations of low-energy localization for constrained devices, \eg{}, XR devices, UAVs, and embedded robotics platforms.

\clearpage

\section*{Acknowledgements}
This publication is part of the grant JDC2024-055088-I, funded by \linebreak {MICIU/AEI/10.13039/501100011033} and the FSE+.

%
%
\bibliographystyle{splncs04}
\bibliography{main.bbl}

\clearpage

\section{Supplementary material}
The full intrinsic matrix can be written as
\begin{equation}\label{eq:full}
    \mat{K} = \begin{bmatrix}
        f_x & \gamma & u_0 \\
        0 & f_y & v_0 \\
        0 & 0 & 1
    \end{bmatrix},
\end{equation}
where the focal length is separated in two components, $f_x$ and~$f_y$. The aspect ratio $\alpha = f_y/f_x\approx 1$, but may differ due to a variety of optical and digital techniques, such as using anamorphic formats, flaws in the sensor manufacturing or digital post-processing. The skew parameter~$\gamma$ corrects for nonrectangular pixels, which also stem from imperfect sensor manufacturing.
The principal point~$(u_0, v_0)$ is the intersection point of the light ray orthogonal to the image plane and passing through the camera center. 

In the main paper, we assumed that the coordinate system is normalized such that the principal point is $(0,0)$, which is often a good approximation if you align it with the center of the image coordinate system. Furthermore, $\alpha=1$ and $\gamma=0$, which is a good approximation for modern cameras. Another reason for reducing the number of unknowns is that the solver becomes faster,
which is essential for real-time computations. Nonlinear refinement can be added in later stages to account for model imperfections.

\subsection{Modified constraints}
We ignore the skew and aspect ratio, \ie{} we assume~$\alpha=1$ and~$\gamma=0$, and modify the governing constraints to account for the full model~\eqref{eq:full}. The affine features are independent of the principal point, and we get
{\footnotesize%
\begin{align} \label{eq:affine-constratins1-mod}
    \frac{a_1}{f} &= \frac{d}{m}(br_{11} - p'_{\text{ref}_1} n_{\text{ref}_1} - \frac{p_{\text{query}_1} - u_0}{f} (br_{31} - p'_{\text{ref}_3} n_{\text{ref}_1} )), \\ \label{eq:affine-constratins2-mod}
    \frac{a_2}{f} &= \frac{d}{m}(br_{12} - p'_{\text{ref}_1} n_{\text{ref}_2} - \frac{p_{\text{query}_1} - u_0}{f} (br_{32} - p'_{\text{ref}_3} n_{\text{ref}_2} )), \\ \label{eq:affine-constratins3-mod}
    \frac{a_3}{f} &= \frac{d}{m}(br_{21} - p'_{\text{ref}_2} n_{\text{ref}_1} - \frac{p_{\text{query}_2} - v_0}{f} (br_{31} - p'_{\text{ref}_3} n_{\text{ref}_1} )), \\ \label{eq:affine-constratins4-mod}
    \frac{a_4}{f} &= \frac{d}{m}(br_{22} - p'_{\text{ref}_2} n_{\text{ref}_2} - \frac{p_{\text{query}_2} - v_0}{f} (br_{32} - p'_{\text{ref}_3} n_{\text{ref}_2} ))\;.
\end{align}%
}%
By inserting~\eqref{eq:affine-constratins1-mod}--\eqref{eq:affine-constratins4-mod} into~\eqref{eq:aff1}--\eqref{eq:aff2},
the following constraints are obtained for scale- and orientation covariant features
{\footnotesize
\begin{equation}
\begin{aligned}\label{eq:sift-const-mod1}
 &d\aref{c}(br_{11} - p'_{\text{ref}_1}  n_{\text{ref}_1} - \frac{p_{\text{query}_1} - u_0}{f} (br_{31} - p'_{\text{ref}_3} n_{\text{ref}_1})) \\
+&d\aref{s}(br_{21} - p'_{\text{ref}_1}  n_{\text{ref}_2} - \frac{p_{\text{query}_1} - u_0}{f} (br_{32} - p'_{\text{ref}_3} n_{\text{ref}_2}))\\
-&\frac{mq\cquery}{f} = 0,
\end{aligned}
\end{equation}
\begin{equation}
\begin{aligned}\label{eq:sift-const-mod2}
 &d\aref{c}(br_{21} - p'_{\text{ref}_2}  n_{\text{ref}_1} - \frac{p_{\text{query}_2} - v_0}{f} (br_{31} - p'_{\text{ref}_3} n_{\text{ref}_1})) \\
+&d\aref{s}(br_{21} - p'_{\text{ref}_2}  n_{\text{ref}_2} - \frac{p_{\text{query}_2} - v_0}{f} (br_{32} - p'_{\text{ref}_3} n_{\text{ref}_2}))\\
-&\frac{mq\squery}{f} = 0\;.
\end{aligned}
\end{equation}
}
Subtracting~\eqref{eq:sift-const-mod1} from~\eqref{eq:sift-const-mod2}, one obtains an orientation-covariant constraint.
{\footnotesize
\begin{equation}
\begin{aligned}\label{eq:ori-const-mod}
    &\aref{c} \squery (b r_{11} - p'_{\text{ref}_1} n_{\text{ref}_1} - \frac{p_{\text{query}_1} - u_0}{f} (b r_{31} - p'_{\text{ref}_3} n_{\text{ref}_1})) +\\
    &\sref \squery (b r_{12} - p'_{\text{ref}_1} n_{\text{ref}_2} - \frac{p_{\text{query}_1} - u_0}{f} (b r_{32} - p'_{\text{ref}_3} n_{\text{ref}_2})) -\\
    &\aref{c} \cquery (b r_{21} - p'_{\text{ref}_2} n_{\text{ref}_1} - \frac{p_{\text{query}_2} - v_0}{f} (b r_{31} - p'_{\text{ref}_3} n_{\text{ref}_1})) -\\
    &\sref \cquery (b r_{22} - p'_{\text{ref}_2} n_{\text{ref}_2} - \frac{p_{\text{query}_2}-v_0}{f} (b r_{32} - p'_{\text{ref}_3} n_{\text{ref}_2})) = 0,
\end{aligned}
\end{equation}
}
Lastly, the principal point affects the point-based constraints
\begin{align} \label{eq:point-projections1-mod}
\frac{p_{\text{query}_1} - u_0}{f} (d\mat{R}_{3,:} \tpref + t_3) - (d\mat{R}_{1,:} \tpref + t_1) &= 0, \\ \label{eq:point-projections2-mod}
\frac{p_{\text{query}_2} - v_0}{f} (d\mat{R}_{3,:} \tpref + t_3) - (d\mat{R}_{2,:} \tpref + t_2) &= 0\;.
\end{align}
These modified constraints need to be multiplied by an appropriate factor to become a polynomial.

\subsection{Comparison to state-of-the-art}
Larsson~\etal{}~\cite{larsson-etal-2018} proposed two solvers including unknown focal length and unknown principal point with and without non-unit aspect ratio, called P4.5Pfuv and P5Pfuva, respectively. These point-based solvers require five 2D-3D correspondences each and do not use the gravity direction.

In our case, we have seven (or eight) degrees of freedom: four extrinsic (1 rotation and 3 translation) and three intrinsic: $f$, $u_0$, $v_0$ (possible with $f_x$ and $f_y$ with unknown aspect ratio). For the 7 DoF problem we require two affine features (UP2PfuvAC), 
two scale- and orientation-covariant features (UP2PfuvSIFT), or three orientation-covariant features (UP3PfuvORI),
to obtain a solution. This reasoning generalizes to the 8 DoF case with non-unit aspect ratio.

\subsection{Analysis of solutions}
We analyze the feasibility of solving the above problems using the computational algebraic geometry tool \texttt{Macaulay2}, which helps us understand if the polynomial system of equations that
arise from the solver configuration admits a solution, and if so, how many solutions there are. The results are shown in~\cref{tab:full-solvers}. A Gröbner basis solver, \eg{},~\cite{larsson-etal-2017}, or resultant-based method can be used to find these solutions.

\begin{table*}[t!]
    \centering
    \caption{Potential solvers. Template size when using~\cite{larsson-etal-2017}.}
    \label{tab:full-solvers}
    \begin{tabular}{lcccccc}
         Solver       & Upright    & DOF & Eqs & Minimal    & Solutions & Template size \\ \hline
         UP2PfuvAC    & \checkmark & 7   & 12 &            & 6   &  $273\times279$ \\
         UP2PfuvSIFT  & \checkmark & 7   & 8  &            & 6   &  $284\times290$ \\
         UP3PfuvORI   & \checkmark & 7   & 9  &            & 6   &  $343\times349$ \\
         UP1PfaAC     & \checkmark & 6   & 6  & \checkmark & TBD & TBD \\
         UP2PfaORI    & \checkmark & 6   & 6  & \checkmark & TBD & TBD \\
         UP2PfuvaAC   & \checkmark & 8   & 12 &            & TBD & TBD \\
         UP2PfuvaSIFT & \checkmark & 8   & 8  & \checkmark & TBD & TBD \\
         UP3PfuvaORI  & \checkmark & 8   & 9  &            & TBD & TBD
    \end{tabular}
\end{table*}

\subsection{Implementation details}
While the Gröbner basis solver~\cite{larsson-etal-2017} can find a solution, it is rarely useful without more intervention. For all cases described above, the equations are linear in~$\tq$, hence the first step used in the main paper can be further applied. \Eg{}, in all cases: UP2PfuvAC, UP2PfuvSIFT, and UP3PfuvORI,
we may write the complete system of equations in a compact form
\begin{equation}
    \mat{M}(r, f, u_0, v_0)\begin{bmatrix}
        \tq \\ 1
    \end{bmatrix}
    =\vec{0},
\end{equation}
where~$\mat{M}$ is a $7\times 4$ matrix, where the coefficients of $m_{ij}$ vary depending on the solver. Again, all subdeterminants of size $4\times 4$ must vanish, but many will be trivially zero. In some cases, it might be beneficial to consider a solution similar to that of \ouro~(in the main paper), where only a subset of the parameters are solved for initially.
The added terms~$u_0$ and $v_0$ are linear, therefore a similar approach to that of the main paper should be possible. This will generate a system of equations in fewer unknowns, upon which a Gröbner basis solver (or similar) can be applied, followed by back-substitution to obtain the remaining unknowns.


\clearpage

\section{Supplementary material}
The full intrinsic matrix can be written as
\begin{equation}\label{eq:full}
    \mat{K} = \begin{bmatrix}
        f_x & \gamma & u_0 \\
        0 & f_y & v_0 \\
        0 & 0 & 1
    \end{bmatrix},
\end{equation}
where the focal length is separated in two components, $f_x$ and~$f_y$. The aspect ratio $\alpha = f_y/f_x\approx 1$, but may differ due to a variety of optical and digital techniques, such as using anamorphic formats, flaws in t he sensor manufacturing or digital post-processing. The skew parameter~$\gamma$ corrects for nonrectangular pixels, which also stem from imperfect sensor manufacturing.
The principal point~$(u_0, v_0)$ is the intersection point of the light ray orthogonal to the image plane and passing through the camera center. 

In the main paper, we assumed that the coordinate system is normalized such that the principal point is $(0,0)$, which is often a good approximation if you align it with the center of the image coordinate system. Furthermore, $\alpha=1$ and $\gamma=0$, which is a good approximation for modern cameras. Another reason for reducing the number of unknowns is that the solver becomes faster,
which is essential for real-time computations. Nonlinear refinement can be added in later stages to account for model imperfections.

\subsection{Modified constraints}
We ignore the skew, and assume~$\gamma=0$, and modify the governing constraints to account for the full model~\eqref{eq:full}. The affine features are independent of the principal point, and we get
{\footnotesize%
\begin{align} \label{eq:affine-constratins1-mod}
    \frac{a_1}{f_x} &= \frac{d}{m}(br_{11} - p'_{\text{ref}_1} n_{\text{ref}_1} - \frac{p_{\text{query}_1}}{f_x} (br_{31} - p'_{\text{ref}_3} n_{\text{ref}_1} )), \\ \label{eq:affine-constratins2-mod}
    \frac{a_2}{f_y} &= \frac{d}{m}(br_{12} - p'_{\text{ref}_1} n_{\text{ref}_2} - \frac{p_{\text{query}_1}}{f_x} (br_{32} - p'_{\text{ref}_3} n_{\text{ref}_2} )), \\ \label{eq:affine-constratins3-mod}
    \frac{a_3}{f_x} &= \frac{d}{m}(br_{21} - p'_{\text{ref}_2} n_{\text{ref}_1} - \frac{p_{\text{query}_2}}{f_y} (br_{31} - p'_{\text{ref}_3} n_{\text{ref}_1} )), \\ \label{eq:affine-constratins4-mod}
    \frac{a_4}{f_y} &= \frac{d}{m}(br_{22} - p'_{\text{ref}_2} n_{\text{ref}_2} - \frac{p_{\text{query}_2}}{f_y} (br_{32} - p'_{\text{ref}_3} n_{\text{ref}_2} ))\;.
\end{align}%
}%
Furthermore, for orientation-covariant features, the following must hold:
{\footnotesize
\begin{equation}
\begin{aligned}\label{eq:ori-const-mod}
    &\aref{c} \squery (b r_{11} - p'_{\text{ref}_1} n_{\text{ref}_1} - \frac{p_{\text{query}_1}}{f_x} (b r_{31} - p'_{\text{ref}_3} n_{\text{ref}_1})) +\\
    &\sref \squery (b r_{12} - p'_{\text{ref}_1} n_{\text{ref}_2} - \frac{p_{\text{query}_1}}{f_y} (b r_{32} - p'_{\text{ref}_3} n_{\text{ref}_2})) -\\
    &\aref{c} \cquery (b r_{21} - p'_{\text{ref}_2} n_{\text{ref}_1} - \frac{p_{\text{query}_2}}{f_x} (b r_{31} - p'_{\text{ref}_3} n_{\text{ref}_1})) -\\
    &\sref \cquery (b r_{22} - p'_{\text{ref}_2} n_{\text{ref}_2} - \frac{p_{\text{query}_2}}{f_y} (b r_{32} - p'_{\text{ref}_3} n_{\text{ref}_2})) = 0,
\end{aligned}
\end{equation}
}
The principal point affects the point-based constraints
\begin{align} \label{eq:point-projections1-mod}
\frac{p_{\text{query}_1} - u_0}{f_x} (d\mat{R}_{3,:} \tpref + t_3) - (d\mat{R}_{1,:} \tpref + t_1) &= 0, \\ \label{eq:point-projections2-mod}
\frac{p_{\text{query}_2} - v_0}{f_y} (d\mat{R}_{3,:} \tpref + t_3) - (d\mat{R}_{2,:} \tpref + t_2) &= 0\;.
\end{align}
These modified constraints need to by multiplied appropriately to become polynomial.

\subsection{Special cases}
\paragraph{Non-unit aspect ratio}
A common case, also used by P4Pf~\cite{kukelova-etal-2016}, is that we separate $f_x$ and $f_y$. In this case, we have six degrees of freedom, and six equations; hence, it is possible to
use a single affine feature to solve the problem (UP1PfaAC).
Similarly, we could use two orientation-covariant features (UP2PfaORI), which also yields six equations, hence is minimal.

\paragraph{Full model}
In this case, we have eight degrees of freedom, and has also been considered by Larsson~\etal{}~(P4.5Pfuv and P5Pfuva) for point-based solvers and without gravity direction, for which the minimal configuration requires five point correspondences. In our case, one would require two affine features (12 equations), however, one could also use only (2+6) constraints, \ie{} the point-based and one affine feature (UP2PfuvaAC).
A minimal case consists of using two scale- and orientation-covariant features (UP2PfuvaSIFT) as each feature yields four equations. Alternatively, three orientation-covariant features could be used (UP3PfuvaORI).

\paragraph{No IMU data}
We could apply this to the case of no known gravity direction as well (unclear if it is interesting for us, and protectable)

\subsection{Analysis of solutions}
We analyze the feasibility of solving the above problems using the computational algebraic geometry tool \texttt{Macaulay2}, which helps us understand if the polynomial systems of equations that
arise from the solver configuration admits a solution, and if so, how many solutions there are, the results are shown in~\cref{tab:full-solvers}.

\begin{table*}[t!]
    \centering
    \caption{Potential solvers.}
    \label{tab:full-solvers}
    \begin{tabular}{lccccl}
         Solver       & Upright    & DOF & Eqs & Minimal    & Solutions \\ \hline
         UP1PfaAC     & \checkmark & 6   & 6  & \checkmark & TBD \\
         UP2PfaORI    & \checkmark & 6   & 6  & \checkmark & TBD \\
         UP2PfuvaAC   & \checkmark & 8   & 12 &            & TBD \\
         UP2PfuvaSIFT & \checkmark & 8   & 8  & \checkmark & TBD \\
         UP3PfuvaORI  & \checkmark & 8   & 9  &            & TBD \\
         P2PfuvaAC    &            & 10  & 12 &            & TBD \\
         P2PfaSIFT    &            & 8   & 8  & \checkmark & TBD \\
         P3PfuvORI    &            & 9   & 9  & \checkmark & TBD \\
         P3PfuvaSIFT  &            & 10  & 12 &            & TBD \\
         P4PfuvaORI   &            & 10  & 12 &            & TBD \\
    \end{tabular}
\end{table*}

\subsection{Implementation details}
For all cases described above, the equations are linear in the translation vector, hence the first step used in the main paper can be further applied. All added terms are linear, hence standard procedure should apply, but at an added computational cost, compared to the proposed solvers in the main paper. A Gröbner basis solver~\cite{larsson-etal-2017} can be used for the reduced system of equations and back-substitution be applied to obtain the complete solution.

\end{document}

%% file: figures/tikzdefinitions.tex
\pgfplotsset{
    p35pf/.style={
        mark=square*, mark options={solid,fill=white}, red, thick, solid
    },
    p4pf/.style={
        mark=square*, mark options={solid,fill=white}, orange, thick, solid
    },
    p1ac/.style={
        mark=square*,mark options={solid,fill=blue}, blue, thick, solid
    },
    up2p/.style={
        mark=*,mark options={solid,fill=white}, purple!80!blue, thick, solid
    },
    up3pf/.style={
        mark=*,mark options={solid,fill=white}, cyan!80!blue, thick, solid
    },
    up1sift/.style={
        mark=*,mark options={solid,fill=pink}, pink, thick, solid
    },
    up1pfac/.style={
        mark=*,mark options={solid,fill=yellow!70!black}, yellow!70!black, thick, dashed
    },
    up2pfori/.style={
        mark=*,mark options={solid,fill=green!80!black}, green!80!black, thick,  dashed
    },
    noise-plot-common/.style={
        width=0.8\textwidth,
        height=0.54\textwidth,
        axis x line*=bottom,
        axis y line*=left,
        ymin = 0,
        xmin = 0,
        label style={font=\Large},
        y tick label style={/pgf/number format/fixed, /pgf/number format/precision=2},
    },
} 

%% file: figures/stability_R.tex
\newcommand{\stabilityrot}[1]{%
\addplot +[#1,no markers] table [x=a, y=b, col sep=comma]%
{data/stability/hist_pnt_0.000000_aff_0.000000_ang_0.000000_scl_0.000000_imu_0.000000_#1_rot.txt};
}

\begin{tikzpicture}[>=latex]
    \begin{axis}[
        width=0.8\textwidth,
        height=0.5\textwidth,
        axis x line*=bottom,
        axis y line*=left,
        ylabel={Frequency} ,
        ymin = 0,
        xmin = -17,
        xmax = 3,
        xtick = {-16, -12, -8, -4, 0},
        xticklabels={$10^{-16}$,$10^{-12}$,$10^{-8}$,$10^{-4}$, 1},
        xlabel={Rotation error},
        label style={font=\Large},
        label style={font=\Large},
    ]
    
    \stabilityrot{p35pf}
    \stabilityrot{p4pf}
    \stabilityrot{p1ac}
    \stabilityrot{up2p}
    \stabilityrot{up1sift}
    \stabilityrot{up3pf}
    \stabilityrot{up1pfac}
    \stabilityrot{up2pfori}
    
\end{axis}
\end{tikzpicture}

%% file: figures/stability_t.tex
\newcommand{\stabilityrot}[1]{%
\addplot +[#1,no markers] table [x=a, y=b, col sep=comma]
{data/stability/hist_pnt_0.000000_aff_0.000000_ang_0.000000_scl_0.000000_imu_0.000000_#1_pos.txt};
}

\begin{tikzpicture}[>=latex]
    \begin{axis}[
        width=0.8\textwidth,
        height=0.5\textwidth,
        axis x line*=bottom,
        axis y line*=left,
        ylabel={Frequency} ,
        ymin = 0,
        xmin = -17,
        xmax = 3,
        xtick = {-16, -12, -8, -4, 0},
        xticklabels={$10^{-16}$,$10^{-12}$,$10^{-8}$,$10^{-4}$, 1},
        xlabel={Position error},
        label style={font=\Large},
    ]
    
    \stabilityrot{p35pf}
    \stabilityrot{p4pf}
    \stabilityrot{p1ac}
    \stabilityrot{up2p}
    \stabilityrot{up1sift}
    \stabilityrot{up3pf}
    \stabilityrot{up1pfac}
    \stabilityrot{up2pfori}
    
\end{axis}
\end{tikzpicture}

%% file: figures/stability_f.tex
\newcommand{\stabilityrot}[1]{%
\addplot +[#1,no markers] table [x=a, y=b, col sep=comma]%
{data/stability/hist_pnt_0.000000_aff_0.000000_ang_0.000000_scl_0.000000_imu_0.000000_#1_focal.txt};
}

\begin{tikzpicture}[>=latex]
    \begin{axis}[
        width=0.8\textwidth,
        height=0.5\textwidth,
        axis x line*=bottom,
        axis y line*=left,
        ylabel={Frequency} ,
        ymin = 0,
        xmin = -17,
        xmax = 3,
        xtick = {-16, -12, -8, -4, 0},
        xticklabels={$10^{-16}$,$10^{-12}$,$10^{-8}$,$10^{-4}$, 1},
        xlabel={Focal length error},
        label style={font=\Large},
    ]
    
    \stabilityrot{p35pf}
    \stabilityrot{p4pf}
    \stabilityrot{up3pf}
    \stabilityrot{up1pfac}
    \stabilityrot{up2pfori}
    
\end{axis}
\end{tikzpicture}

%% file: figures/legend_no_markers.tex
\newenvironment{customlegend}[1][]{%
    \begingroup
    \csname pgfplots@init@cleared@structures\endcsname
    \pgfplotsset{#1}%
}{%
    \csname pgfplots@createlegend\endcsname
    \endgroup
}%
\def\addlegendimage{\csname pgfplots@addlegendimage\endcsname}

\begin{tikzpicture}[>=latex]
\begin{customlegend}[legend columns=8,legend style={at={(7.8,-1.2)},align=left,column sep=2ex},
        legend entries={
            P1AC~\cite{ventura-etal-iccv-2023},
            UP2P~\cite{kukelova-etal-2010},
            UP1SIFT~\cite{ventura-etal-cvpr-2024},
            P3.5Pf~\cite{larsson-etal-2017},
            P4Pf~\cite{kukelova-etal-2016},
            UP2.5Pf~\cite{kukelova-etal-2010},
            UP1PfAC (Our),
            UP2PfORI (Our)
        }]
        \addlegendimage{p1ac,no markers}
        \addlegendimage{up2p,no markers}
        \addlegendimage{up1sift,no markers}
        \addlegendimage{p35pf,no markers}
        \addlegendimage{p4pf,no markers}
        \addlegendimage{up3pf,no markers}
        \addlegendimage{up1pfac,no markers}
        \addlegendimage{up2pfori,no markers}
        \end{customlegend}
\end{tikzpicture}

%% file: figures/noise_point_R.tex
\newcommand{\noiserot}[1]{%
\addplot +[#1] table [x=a, y=b, col sep=comma]%
{data/noise/noise_pnt_#1_rot.csv};
}

\begin{tikzpicture}[>=latex]
    \begin{axis}[
        noise-plot-common,
        ylabel={Rotation error} ,
        xmax = 2,
        xlabel={Point noise (px.)},
    ]
    
    \noiserot{p35pf}
    \noiserot{p4pf}
    \noiserot{p1ac}
    \noiserot{up2p}
    \noiserot{up1sift}
    \noiserot{up3pf}
    \noiserot{up1pfac}
    \noiserot{up2pfori}
    
\end{axis}
\end{tikzpicture}

%% file: figures/noise_point_t.tex
\newcommand{\noiserot}[1]{%
\addplot +[#1] table [x=a, y=b, col sep=comma]%
{data/noise/noise_pnt_#1_pos.csv};
}

\begin{tikzpicture}[>=latex]
    \begin{axis}[
        noise-plot-common,
        ylabel={Position error} ,
        xmax = 2,
        xlabel={Point noise (px.)},
    ]
    
    \noiserot{p35pf}
    \noiserot{p4pf}
    \noiserot{p1ac}
    \noiserot{up2p}
    \noiserot{up1sift}
    \noiserot{up3pf}
    \noiserot{up1pfac}
    \noiserot{up2pfori}
    
\end{axis}
\end{tikzpicture}

%% file: figures/noise_point_f.tex
\newcommand{\noiserot}[1]{%
\addplot +[#1] table [x=a, y=b, col sep=comma]%
{data/noise/noise_pnt_#1_focal.csv};
}

\begin{tikzpicture}[>=latex]
    \begin{axis}[
        noise-plot-common,
        ylabel={Focal length error} ,
        xmax = 2,
        xlabel={Point noise (px.)},
    ]
    
    \noiserot{p35pf}
    \noiserot{p4pf}
    \noiserot{up3pf}
    \noiserot{up1pfac}
    \noiserot{up2pfori}
    
\end{axis}
\end{tikzpicture}

%% file: figures/noise_normal_R.tex
\newcommand{\noiserot}[1]{%
\addplot +[#1] table [x=a, y=b, col sep=comma]%
{data/noise/noise_normal_#1_rot.csv};
}

\begin{tikzpicture}[>=latex]
    \begin{axis}[
        noise-plot-common,
        ylabel={Rotation error} ,
        xmax = 1,
        xlabel={Normal noise (deg.)},
    ]
    
    \noiserot{p35pf}
    \noiserot{p4pf}
    \noiserot{p1ac}
    \noiserot{up2p}
    \noiserot{up1sift}
    \noiserot{up3pf}
    \noiserot{up1pfac}
    \noiserot{up2pfori}
    
\end{axis}
\end{tikzpicture}

%% file: figures/noise_normal_t.tex
\newcommand{\noiserot}[1]{%
\addplot +[#1] table [x=a, y=b, col sep=comma]%
{data/noise/noise_normal_#1_pos.csv};
}

\begin{tikzpicture}[>=latex]
    \begin{axis}[
        noise-plot-common,
        ylabel={Position error} ,
        xmax = 1,
        xlabel={Normal noise (deg.)},
    ]
    
    \noiserot{p35pf}
    \noiserot{p4pf}
    \noiserot{p1ac}
    \noiserot{up2p}
    \noiserot{up1sift}
    \noiserot{up3pf}
    \noiserot{up1pfac}
    \noiserot{up2pfori}
    
\end{axis}
\end{tikzpicture}

%% file: figures/noise_normal_f.tex
\newcommand{\noiserot}[1]{%
\addplot +[#1] table [x=a, y=b, col sep=comma]%
{data/noise/noise_normal_#1_focal.csv};
}

\begin{tikzpicture}[>=latex]
    \begin{axis}[
        noise-plot-common,
        ylabel={Focal length error} ,
        xmax = 1,
        xlabel={Normal noise (deg.)},
    ]
    
    \noiserot{p35pf}
    \noiserot{p4pf}
    \noiserot{up3pf}
    \noiserot{up1pfac}
    \noiserot{up2pfori}
    
\end{axis}
\end{tikzpicture}

%% file: figures/noise_angle_R.tex
\newcommand{\noiserot}[1]{%
\addplot +[#1] table [x=a, y=b, col sep=comma]%
{data/noise/noise_ang_#1_rot.csv};
}

\begin{tikzpicture}[>=latex]
    \begin{axis}[
        noise-plot-common,
        ylabel={Rotation error} ,
        xmax = 1.0,
        xlabel={Orientation noise (deg.)},
    ]
    
    \noiserot{p35pf}
    \noiserot{p4pf}
    \noiserot{p1ac}
    \noiserot{up2p}
    \noiserot{up1sift}
    \noiserot{up3pf}
    \noiserot{up1pfac}
    \noiserot{up2pfori}
    
\end{axis}
\end{tikzpicture}

%% file: figures/noise_angle_t.tex
\newcommand{\noiserot}[1]{%
\addplot +[#1] table [x=a, y=b, col sep=comma]%
{data/noise/noise_ang_#1_pos.csv};
}

\begin{tikzpicture}[>=latex]
    \begin{axis}[
        noise-plot-common,
        ylabel={Position error} ,
        xmax = 1,
        xlabel={Orientation noise (deg.)},
    ]
    
    \noiserot{p35pf}
    \noiserot{p4pf}
    \noiserot{p1ac}
    \noiserot{up2p}
    \noiserot{up1sift}
    \noiserot{up3pf}
    \noiserot{up1pfac}
    \noiserot{up2pfori}
    
\end{axis}
\end{tikzpicture}

%% file: figures/noise_angle_f.tex
\newcommand{\noiserot}[1]{%
\addplot +[#1] table [x=a, y=b, col sep=comma]%
{data/noise/noise_ang_#1_focal.csv};
}

\begin{tikzpicture}[>=latex]
    \begin{axis}[
        noise-plot-common,
        ylabel={Focal length error} ,
        xmax = 1,
        xlabel={Orientation noise (deg.)},
    ]
    
    \noiserot{p35pf}
    \noiserot{p4pf}
    \noiserot{up3pf}
    \noiserot{up1pfac}
    \noiserot{up2pfori}
    
\end{axis}
\end{tikzpicture}

%% file: figures/noise_aff_R.tex
\newcommand{\noiserot}[1]{%
\addplot +[#1] table [x=a, y=b, col sep=comma]%
{data/noise/noise_aff_#1_rot.csv};
}

\begin{tikzpicture}[>=latex]
    \begin{axis}[
        noise-plot-common,
        ylabel={Rotation error} ,
        xmax = 0.1,
        xlabel={Affine noise},
    ]
    
    \noiserot{p35pf}
    \noiserot{p4pf}
    \noiserot{p1ac}
    \noiserot{up2p}
    \noiserot{up1sift}
    \noiserot{up3pf}
    \noiserot{up1pfac}
    \noiserot{up2pfori}
    
\end{axis}
\end{tikzpicture}

%% file: figures/noise_aff_t.tex
\newcommand{\noiserot}[1]{%
\addplot +[#1] table [x=a, y=b, col sep=comma]%
{data/noise/noise_aff_#1_pos.csv};
}

\begin{tikzpicture}[>=latex]
    \begin{axis}[
        noise-plot-common,
        ylabel={Position error} ,
        xmax = 0.1,
        xlabel={Affine noise},
    ]
    
    \noiserot{p35pf}
    \noiserot{p4pf}
    \noiserot{p1ac}
    \noiserot{up2p}
    \noiserot{up1sift}
    \noiserot{up3pf}
    \noiserot{up1pfac}
    \noiserot{up2pfori}
    
\end{axis}
\end{tikzpicture}

%% file: figures/noise_aff_f.tex
\newcommand{\noiserot}[1]{%
\addplot +[#1] table [x=a, y=b, col sep=comma]%
{data/noise/noise_aff_#1_focal.csv};
}

\begin{tikzpicture}[>=latex]
    \begin{axis}[
        noise-plot-common,
        ylabel={Focal length error} ,
        xmax = 0.1,
        xlabel={Affine noise},
    ]
    
    \noiserot{p35pf}
    \noiserot{p4pf}
    \noiserot{up3pf}
    \noiserot{up1pfac}
    \noiserot{up2pfori}
    
\end{axis}
\end{tikzpicture}

%% file: figures/noise_imu_R.tex
\newcommand{\noiserot}[1]{%
\addplot +[#1] table [x=a, y=b, col sep=comma]%
{data/noise/noise_imu_#1_rot.csv};
}

\begin{tikzpicture}[>=latex]
    \begin{axis}[
        noise-plot-common,
        ylabel={Rotation error} ,
        xmax = 0.5,
        xlabel={IMU noise (deg.)},
    ]
    
    \noiserot{p35pf}
    \noiserot{p4pf}
    \noiserot{p1ac}
    \noiserot{up2p}
    \noiserot{up1sift}
    \noiserot{up3pf}
    \noiserot{up1pfac}
    \noiserot{up2pfori}
    
\end{axis}
\end{tikzpicture}

%% file: figures/noise_imu_t.tex
\newcommand{\noiserot}[1]{%
\addplot +[#1] table [x=a, y=b, col sep=comma]%
{data/noise/noise_imu_#1_pos.csv};
}

\begin{tikzpicture}[>=latex]
    \begin{axis}[
        noise-plot-common,
        ylabel={Position error} ,
        xmax = 0.5,
        xlabel={IMU noise (deg.)},
    ]
    
    \noiserot{p35pf}
    \noiserot{p4pf}
    \noiserot{p1ac}
    \noiserot{up2p}
    \noiserot{up1sift}
    \noiserot{up3pf}
    \noiserot{up1pfac}
    \noiserot{up2pfori}
    
\end{axis}
\end{tikzpicture}

%% file: figures/noise_imu_f.tex
\newcommand{\noiserot}[1]{%
\addplot +[#1] table [x=a, y=b, col sep=comma]%
{data/noise/noise_imu_#1_focal.csv};
}

\begin{tikzpicture}[>=latex]
    \begin{axis}[
        noise-plot-common,
        ylabel={Focal length error} ,
        xmax = 0.5,
        xlabel={IMU noise (deg.)},
    ]
    
    \noiserot{p35pf}
    \noiserot{p4pf}
    \noiserot{up3pf}
    \noiserot{up1pfac}
    \noiserot{up2pfori}
    
\end{axis}
\end{tikzpicture}

%% file: figures/legend.tex
\newenvironment{customlegend}[1][]{%
    \begingroup
    \csname pgfplots@init@cleared@structures\endcsname
    \pgfplotsset{#1}%
}{%
    \csname pgfplots@createlegend\endcsname
    \endgroup
}%
\def\addlegendimage{\csname pgfplots@addlegendimage\endcsname}

\begin{tikzpicture}[>=latex]
\begin{customlegend}[legend columns=8,legend style={at={(7.8,-1.2)},align=left,column sep=2ex},
        legend entries={
            P1AC~\cite{ventura-etal-iccv-2023},
            UP2P~\cite{kukelova-etal-2010},
            UP1SIFT~\cite{ventura-etal-cvpr-2024},
            P3.5Pf~\cite{larsson-etal-2017},
            P4Pf~\cite{kukelova-etal-2016},
            UP2.5Pf~\cite{kukelova-etal-2010},
            UP1PfAC (Our),
            UP2PfORI (Our)
        }]
        \addlegendimage{p1ac}
        \addlegendimage{up2p}
        \addlegendimage{up1sift}
        \addlegendimage{p35pf}
        \addlegendimage{p4pf}
        \addlegendimage{up3pf}
        \addlegendimage{up1pfac}
        \addlegendimage{up2pfori}
        \end{customlegend}
\end{tikzpicture}

%% file: figures/ransac1.tex
\newcommand{\noiserot}[1]{%
\addplot +[#1, no markers] table [x=a, y=b, col sep=comma]%
{data/ransac1/ransac_iter_#1.csv};
}

\begin{tikzpicture}[>=latex]
    \begin{axis}[
        width=0.8\textwidth,
        height=0.5\textwidth,
        axis x line*=bottom,
        axis y line*=left,
        ylabel={Number of inliers} ,
        ymin = 0,
        xmin = 0,
        xmax = 1,
        xlabel={Time (ms)},
        label style={font=\Large},
    ]
    
    \noiserot{p35pf}
    \noiserot{p4pf}
    \noiserot{up3pf}
    \noiserot{up1pfac}
    \noiserot{up2pfori}
    
\end{axis}
\end{tikzpicture}

%% file: figures/ransac2.tex
\newcommand{\noiserot}[1]{%
\addplot +[#1, no markers] table [x=a, y=b, col sep=comma]%
{data/ransac2/ransac_outliers_#1.csv};
}

\begin{tikzpicture}[>=latex]
    \begin{axis}[
        width=0.8\textwidth,
        height=0.5\textwidth,
        axis x line*=bottom,
        axis y line*=left,
        ylabel={Execution time (ms)} ,
        ymin = 0,
        xmin = 0.1,
        xmax = 0.75,
        xlabel={Outlier ratio},
        label style={font=\Large},
    ]
    
    \noiserot{p35pf}
    \noiserot{p4pf}
    \noiserot{up3pf}
    \noiserot{up1pfac}
    \noiserot{up2pfori}
    
\end{axis}
\end{tikzpicture}

%% file: figures/legend_focal.tex
\newenvironment{customlegend}[1][]{%
    \begingroup
    \csname pgfplots@init@cleared@structures\endcsname
    \pgfplotsset{#1}%
}{%
    \csname pgfplots@createlegend\endcsname
    \endgroup
}%
\def\addlegendimage{\csname pgfplots@addlegendimage\endcsname}

\begin{tikzpicture}[>=latex]
\begin{customlegend}[legend columns=5,legend style={at={(7.8,-1.2)},align=left,column sep=2ex},
        legend entries={
            P3.5Pf~\cite{larsson-etal-2017},
            P4Pf~\cite{kukelova-etal-2016},
            UP2.5Pf~\cite{kukelova-etal-2010},
            UP1PfAC (Our),
            UP2PfORI (Our)
        }]
        \addlegendimage{p35pf,no markers}
        \addlegendimage{p4pf,no markers}
        \addlegendimage{up3pf,no markers}
        \addlegendimage{up1pfac,no markers}
        \addlegendimage{up2pfori,no markers}
        \end{customlegend}
\end{tikzpicture}

%% file: tables_cambridge.tex
\begin{table}[htbp!]
\caption{\textbf{Cambridge Landmarks} Median errors in position, rotation and focal length, and median times for the tested solvers in GC-RANSAC, using SP+LG and S3Esti, best method highlighted differently for \textit{calib} and \textbf{semi-calib} methods.
\vspace{-8px}}
\label{tab:cambridge_sp}
\resizebox{\textwidth}{!}{
\begin{tabular}{ll|cccc|cccc|cccc|cccc|cccc}
\toprule
 & & \multicolumn{4}{c|}{GreatCourt} & \multicolumn{4}{c|}{KingsCollege} & \multicolumn{4}{c|}{OldHospital} & \multicolumn{4}{c|}{ShopFacade} & \multicolumn{4}{c}{StMarysChurch} \\
 & & cm & $^\circ$ & $f_{\text{err}}$ & ms& cm & $^\circ$ & $f_{\text{err}}$ & ms& cm & $^\circ$ & $f_{\text{err}}$ & ms& cm & $^\circ$ & $f_{\text{err}}$ & ms& cm & $^\circ$ & $f_{\text{err}}$ & ms\\
\midrule
\midrule
\multirow{3}{*}[-0.0ex]{\rotatebox{90}{\scriptsize Known $f$}} & P1AC~\cite{ventura-etal-iccv-2023} & 32.6 & 0.16 & - & 23.3 & 16.6 & 0.28 & - & \textit{32.4} & 21.2 & 0.36 & - & 35.2 & 6.7 & 0.32 & - & 36.8 & 11.2 & 0.35 & - & 34.3 \\
& UP2P~\cite{kukelova-etal-2010} & \textit{31.6} & \textit{0.15} & - & 23.1 & \textit{16.1} & \textit{0.28} & - & 33.4 & \textit{21.1} & \textit{0.35} & - & 31.8 & \textit{6.5} & \textit{0.30} & - & 34.1 & \textit{10.8} & \textit{0.33} & - & 33.9 \\
& UP1SIFT~\cite{ventura-etal-cvpr-2024} & 45.9 & 0.20 & - & \textit{22.5} & 20.4 & 0.32 & - & 34.0 & 44.5 & 0.64 & - & \textit{31.2} & 8.4 & 0.38 & - & \textit{33.5} & 17.0 & 0.55 & - & \textit{29.1} \\
\cline{2-22}
\multirow{5}{*}[-0.7ex]{\rotatebox{90}{{\scriptsize Unknown $f$}}}  & P4Pf~\cite{kukelova-etal-2016} & 61.5 & 0.14 & 0.010 & 26.9 & 36.3 & 0.30 & \textbf{0.011} & 48.3 & 58.7 & 0.42 & 0.015 & 40.9 & 14.6 & 0.30 & 0.011 & 39.9 & 23.9 & 0.32 & 0.014 & 42.3 \\
& P3.5Pf~\cite{larsson-etal-2017} & 61.3 & 0.14 & 0.010 & 28.0 & 36.0 & 0.30 & \textbf{0.011} & 45.2 & 58.0 & 0.42 & 0.015 & 38.6 & 14.6 & 0.30 & 0.011 & 41.4 & 23.9 & 0.32 & 0.014 & 41.2 \\
& UP2.5Pf~\cite{kukelova-etal-2010} & 61.6 & 0.14 & 0.010 & 23.8 & 35.9 & 0.29 & \textbf{0.011} & 41.1 & 57.2 & 0.42 & 0.015 & 38.0 & 14.4 & 0.30 & 0.011 & 35.4 & 23.7 & 0.32 & \textbf{0.013} & 37.0 \\
& \oura & \textbf{57.7} & \textbf{0.13} & \textbf{0.009} & \textbf{22.6} & 35.6 & \textbf{0.28} & 0.012 & \textbf{39.9} & 52.3 & 0.38 & \textbf{0.013} & \textbf{34.2} & \textbf{10.6} & \textbf{0.26} & \textbf{0.008} & \textbf{33.9} & \textbf{22.6} & \textbf{0.31} & 0.014 & \textbf{34.3} \\
& \ouro & 57.9 & \textbf{0.13} & \textbf{0.009} & 24.0 & \textbf{35.3} & 0.28 & 0.012 & 41.7 & \textbf{52.2} & \textbf{0.37} & 0.014 & 37.1 & 10.9 & \textbf{0.26} & \textbf{0.008} & 34.3 & 22.9 & \textbf{0.31} & 0.014 & 38.2 \\
\bottomrule
\bottomrule
\end{tabular}
}
\end{table}

\begin{table}[htbp!]
\caption{\textbf{Cambridge Landmarks} Median errors in position, rotation and focal length, and median times  the tested solvers in GC-RANSAC, using SIFT and Nearest Neighbors, best method highlighted differently for \textit{calib} and \textbf{semi-calib} methods.
\vspace{-8px}}
\label{tab:cambridge_sift}
\resizebox{\textwidth}{!}{
\begin{tabular}{ll|cccc|cccc|cccc|cccc|cccc}
\toprule
 & & \multicolumn{4}{c|}{GreatCourt} & \multicolumn{4}{c|}{KingsCollege} & \multicolumn{4}{c|}{OldHospital} & \multicolumn{4}{c|}{ShopFacade} & \multicolumn{4}{c}{StMarysChurch} \\
 & & cm & $^\circ$ & $f_{\text{err}}$ & ms& cm & $^\circ$ & $f_{\text{err}}$ & ms& cm & $^\circ$ & $f_{\text{err}}$ & ms& cm & $^\circ$ & $f_{\text{err}}$ & ms& cm & $^\circ$ & $f_{\text{err}}$ & ms\\
\midrule
\midrule
\multirow{3}{*}[-0.0ex]{\rotatebox{90}{\scriptsize Known $f$}} & P1AC~\cite{ventura-etal-iccv-2023} & 59.5 & 0.21 & - & \textit{15.2} & 17.8 & \textit{0.28} & - & \textit{34.5} & 26.6 & 0.45 & - & 20.1 & 8.1 & 0.30 & - & 21.1 & 14.9 & 0.42 & - & 22.4 \\
& UP2P~\cite{kukelova-etal-2010} & \textit{55.1} & \textit{0.19} & - & 16.0 & \textit{17.4} & \textit{0.28} & - & 34.8 & \textit{25.1} & \textit{0.42} & - & 20.5 & \textit{6.8} & \textit{0.28} & - & 20.8 & \textit{14.8} & \textit{0.40} & - & 22.5 \\
& UP1SIFT~\cite{ventura-etal-cvpr-2024} & 58.4 & 0.20 & - & 15.4 & 18.6 & 0.29 & - & 36.3 & 29.4 & 0.48 & - & \textit{18.9} & 7.6 & 0.30 & - & \textit{20.3} & 15.4 & 0.42 & - & \textit{21.3} \\
\cline{2-22}
\multirow{5}{*}[-0.7ex]{\rotatebox{90}{{\scriptsize Unknown $f$}}} & P4Pf~\cite{kukelova-etal-2016} & 75.8 & 0.16 & \textbf{0.011} & 23.3 & 37.9 & 0.31 & \textbf{0.012} & 41.9 & 51.1 & 0.39 & 0.017 & 25.9 & 11.2 & 0.28 & \textbf{0.008} & 23.6 & 28.7 & \textbf{0.37} & 0.018 & 28.4 \\
& P3.5Pf~\cite{larsson-etal-2017} & 75.8 & 0.16 & \textbf{0.011} & 22.7 & 38.0 & 0.32 & \textbf{0.012} & 42.7 & 51.7 & 0.39 & 0.017 & 27.8 & 11.2 & 0.28 & \textbf{0.008} & 25.8 & 28.6 & \textbf{0.37} & 0.018 & 31.2 \\
& UP3Pf~\cite{kukelova-etal-2010} & 75.6 & 0.16 & \textbf{0.011} & 20.5 & 37.8 & 0.31 & \textbf{0.012} & 40.1 & 51.9 & 0.39 & \textbf{0.017} & 24.2 & 11.2 & 0.28 & \textbf{0.008} & \textbf{20.6} & 28.4 & 0.38 & 0.018 & 26.4 \\
& \oura & \textbf{70.5} & \textbf{0.15} & 0.012 & \textbf{16.7} & \textbf{37.6} & \textbf{0.30} & \textbf{0.012} & 35.5 & \textbf{48.4} & \textbf{0.33} & 0.018 & \textbf{21.0} & 11.1 & \textbf{0.27} & \textbf{0.008} & 20.8 & \textbf{27.7} & 0.38 & \textbf{0.017} & \textbf{24.2} \\
& \ouro & 70.8 & \textbf{0.15} & \textbf{0.011} & 17.7 & 37.8 & \textbf{0.30} & \textbf{0.012} & \textbf{35.4} & 50.4 & 0.36 & 0.018 & 23.8 & \textbf{11.0} & \textbf{0.27} & \textbf{0.008} & 21.6 & 28.1 & 0.38 & 0.018 & 27.1 \\
\bottomrule
\end{tabular}
}
\end{table}

%% file: table_aachen.tex
\begin{table}
\vspace{-10px}
\caption{\textbf{Aachen Dataset} Performance for various solvers and features using GC-RANSAC, with percentage error recalls for thresholds 0.25m/2$^\circ$, 0.5m/5$^\circ$, 5.0m/10$^\circ$ and $f_{\text{err}}$. Best method highlighted differently for \textit{calib} and \textbf{semi-calib} methods.
\vspace{-8px}}
\label{tab:aachen_sp}
\centering
\resizebox{0.77\textwidth}{!}{

\begin{tabular}{ll|ccc|ccc|cc||ccc|ccc|cc}
\toprule
 & & \multicolumn{8}{c||}{SuperPoint+LightGlue+S3Esti} & \multicolumn{8}{c}{SIFT+Nearest Neighbors} \\
 & & \multicolumn{3}{c|}{Day} & \multicolumn{3}{c|}{Night} & $f_{\text{err}}$ & ms  & \multicolumn{3}{c|}{Day} & \multicolumn{3}{c|}{Night} & $f_{\text{err}}$  & ms \\
 
\midrule
\midrule
\multirow{3}{*}[-0.0ex]{\rotatebox{90}{\scriptsize Known $f$}} & P1AC~\cite{ventura-etal-iccv-2023} & 69.5 & 89.6 & 99.2 & 70.7 & \textit{88.0} & \textit{99.0} & - & 15.7 & 59.8 & 76.3 & 87.7 & \textit{21.5} & \textit{25.7} & \textit{27.7} & - & 29.1 \\
& UP2P~\cite{kukelova-etal-2010} & \textit{71.8} & \textit{90.0} & \textit{99.3} & \textit{71.2} & 87.4 & 98.4 & - & \textit{15.6} & \textit{61.3} & \textit{76.9} & \textit{87.9} & \textit{21.5} & \textit{25.7} & \textit{27.7} & - & 26.9 \\
& UP1SIFT~\cite{ventura-etal-cvpr-2024} & 60.9 & 84.2 & 93.8 & 59.7 & 79.6 & 92.7 & - & 16.4 & 59.0 & 75.0 & 87.0 & 19.4 & 23.0 & 26.2 & - & \textit{24.8} \\
\midrule
\multirow{5}{*}[-0.7ex]{\rotatebox{90}{{\scriptsize Unknown $f$}}} & P4Pf~\cite{kukelova-etal-2016} & 46.1 & 70.0 & 96.8 & 54.5 & 74.9 & 95.8 & 0.010 & \textbf{16.7} & 37.4 & 57.0 & 85.9 & \textbf{19.4} & 20.9 & 26.2 & 0.019 & 37.9 \\
& P3.5Pf~\cite{larsson-etal-2017} & 45.6 & 69.7 & 96.7 & 54.5 & 74.9 & 95.3 & 0.010 & 20.3  & 37.7 & 57.8 & 85.7 & \textbf{19.4} & 21.5 & 26.2 & 0.018 & 34.8 \\
& UP2.5Pf~\cite{kukelova-etal-2010} & 46.5 & 68.8 & 96.6 & 55.0 & 73.8 & 95.3 & 0.010 & 18.6 & 37.9 & 57.4 & 85.4 & 18.8 & 21.5 & 26.2 & 0.018 & 32.2\\
& \oura & \textbf{48.5} & 70.8 & 97.9 & 60.2 & \textbf{80.1} & \textbf{97.4} & \textbf{0.009} & \textbf{16.7} & 38.8 & \textbf{58.6} & \textbf{86.3} & 18.3 & 22.0 & \textbf{27.2} & \textbf{0.017} & \textbf{28.0} \\
& \ouro & 48.4 & \textbf{71.2} & \textbf{98.2} & \textbf{62.3} & \textbf{80.1} & \textbf{97.4} & \textbf{0.009} & 19.8 & \textbf{39.7} & 58.4 & \textbf{86.3} & \textbf{19.4} & \textbf{22.5} & 26.7 & \textbf{0.017} & 30.8 \\
\bottomrule

\end{tabular}
}
\end{table}